\documentclass[twocolumn]{article}

\usepackage{geometry}
\usepackage{authblk}
\usepackage{cite}
\usepackage[utf8]{inputenc}
\usepackage{multirow}
\usepackage{graphicx}
\usepackage{outlines}
\usepackage[dvipsnames]{xcolor}
\usepackage{colortbl}
\usepackage{textcomp}
\usepackage{float}
\usepackage[normalem]{ulem}
\usepackage{comment}
\usepackage{amssymb}
\usepackage{mathtools}
\usepackage{booktabs}
\usepackage{tikz}
\usepackage{color}
\usepackage[titles]{tocloft}
\usepackage[mathscr]{eucal}

\definecolor{greyblue}{rgb}{0.5255,0.6039,0.6902}

\definecolor{lightgreyblue}{rgb}{0.6314,0.7373,0.7843}
\definecolor{lightgreygreen}{rgb}{0.6392,0.7255,0.6627}
\definecolor{lightgreyred}{rgb}{0.6824,0.6471,0.6863}
\definecolor{lightgreyyellow}{rgb}{0.8078,0.7529,0.6588}
\definecolor{lightgrey}{gray}{0.9}
\definecolor{lightestgrey}{gray}{0.95}

\newcommand{\onlinecite}[1]{\hspace{-1 ex} \nocite{#1}\citenum{#1}} 

\let\OLDthebibliography\thebibliography
\renewcommand\thebibliography[1]{
  \OLDthebibliography{#1}
  \setlength{\parskip}{0pt}
  \setlength{\itemsep}{0pt plus 0.3ex}
}
  
\title{Relating Superconducting Optoelectronic Networks to Classical Neurodynamics}
\author[1]{\Large{Jeffrey M. Shainline$^{*}$, Bryce A. Primavera, and Ryan O'Loughlin}
\\
\textit{\large{National Institute of Standards and Technology}}
\\
\textit{\large{325 Broadway, Boulder, CO, USA, 80305}}
\\
\vspace{0.5em}
\small{$^*$jeffrey.shainline@nist.gov}
}
\date{\today}%

\begin{document}

\twocolumn[
\begin{@twocolumnfalse}
\maketitle
\begin{abstract}
The circuits comprising superconducting optoelectronic synapses, dendrites, and neurons are described by numerically cumbersome and formally opaque coupled differential equations. Reference \onlinecite{shainline2023phenomenological} showed that a phenomenological model of superconducting loop neurons eliminates the need to solve the Josephson circuit equations that describe synapses and dendrites. The initial goal of the model was to decrease the time required for simulations, yet an additional benefit of the model was increased transparency of the underlying neural circuit operations and conceptual clarity regarding the connection of loop neurons to spin systems, pulse-coupled oscillators, and biological spiking neurons. Whereas the original model simplified the treatment of the Josephson-junction dynamics, essentially by only considering low-pass versions of the dendritic outputs, the model resorted to an awkward treatment of spikes generated by semiconductor transmitter circuits that required explicitly checking for threshold crossings and distinct treatment of time steps wherein somatic threshold is reached. Here we extend that model to simplify the treatment of spikes coming from somas, again making use of the fact that in neural systems the downstream recipients of spike events almost always perform low-pass filtering. The phenomenological model is derived from a Lagrangian, providing a formalism for superconducting optoelectronic networks on the foundation of least-action principles. We provide comparisons between the first and second phenomenological models, quantifying the accuracy of the additional approximations. We identify regions of circuit parameter space in which the extended model works well and regions where it works poorly. For some circuit parameters it is possible to represent the downstream dendritic response to a single spike as well as coincidences or sequences of spikes, indicating the model is not simply a reduction to rate coding. A simple algorithm is given for numerical analysis amenable to parallelized implementation on graphics or tensor processing units. The governing equations are shown to be nearly identical to those ubiquitous in the neuroscience literature for modeling leaky-integrator dendrites and neurons. This formal similarity and the overall transparency of the model provide a concise framework in which to leverage substantial prior work in algorithm development, including artificial neural networks and Hopfield models.

\vspace{3em}
\end{abstract}
\end{@twocolumnfalse}
]

\section{\label{sec:introduction}Introduction}
A promising route to better understand the brain and to leverage its principles to perform useful computations is to construct technological systems that embody neural principles. This field of research is known as neuromorphic computing \cite{nawrocki2016mini, furber2016large, schuman2017survey}. Within this domain, a specific approach is being pursued that aims to leverage light for spike-based communication at the single-photon level in conjunction with superconducting circuitry to perform fast, energy efficient computation \cite{shainline2018circuit, shainline2021optoelectronic}. Such systems are referred to as superconducting optoelectronic networks (SOENs). Much progress toward realization of these systems has been made in hardware \cite{buckley2017all, chiles2017multi, chiles2018design, mccaughan2019superconducting,  khan2022superconducting, primavera2023programmable}, but to make them a useful technology, appreciable theoretical development and modeling is necessary. For theoretical advances to be possible, an accurate, numerically efficient, and conceptually transparent formalism is required.

The most physically realistic models of biological neurons are based on the Hodgkin-Huxley equations, a system of four coupled, first-order, ordinary differential equations that capture the dynamics of four types of ion channels with voltage-dependent conductances through the cell membrane \cite{hodgkin1952quantitative,gerstner2002spiking}. Similarly, the microscopic components that constitute SOENs are based on superconducting quantum interference devices (SQUIDs, \cite{van1998principles, kadin1999introduction, tinkham2004introduction, clarke2006squid}), in which the interplay between two Josephson junctions (JJs) and an output low-pass filter can be expressed as a system of five first-order equations \cite{shainline2023phenomenological}. The correspondence between the superconducting and biological systems is remarkable \cite{crotty2010josephson}. In computational neuroscience, the full Hodgkin-Huxley equations are not commonly used to represent large circuits and networks due to the numerical overhead. Instead, simplified models that capture the relevant functionality with more efficient numerical implementation are used in practice with a great deal of success in explaining measured data. This type of phenomenological model was developed for SOENs in Ref.\,\onlinecite{shainline2023phenomenological} to strike a compromise between accuracy and numerical speed. That model reduced the problem to a single, first-order, ordinary differential equation representing each synapse, dendrite, or soma.

In addition to numerical speed, phenomenological models can also aid in the interpretability of behavior. The complexity of four interacting, voltage-dependent ion channels can be simplified to a single leaky integrator equation. In neuroscience, the communication of action potentials between neurons is another example of complicated behavior that is often not explicitly treated in models, particularly models that are intended to scale to very large numbers of interacting elements. In many cases, the discrete and discontinuous behavior of spikes is important, such as for detecting sharp transitions or coincidences, but in other cases the encumberance of calculating and recording spike times is not worth the added insight, and the added insight may even be counterproductive, like tracking the positions and momenta of all atoms in a gas when thermodynamic quantities will suffice. The phenomenological model of Ref.\,\onlinecite{shainline2023phenomenological} greatly simplified the analysis of the computational aspects of synapses, dendrites, and somas, but the model explicitly included spikes as an essential mechanism of interneuronal communication. In that sense, the model made it only halfway to the goal of simplifying the formalism.

Here we extend that framework by eliminating the need to model spikes. We do not claim that spikes are irrelevant to the computations or communication protocols implemented by neurons, but rather that a model without spikes can provide a helpful treatment of a system that can serve as a foundation for development of algorithms and elucidation of connections to neuroscience and other complex systems. In the simplest theoretical treatments of neural systems, a neuron's activity is reduced to a rate. In one treatment of such a model \cite{hopfield1986computing}, Hopfield and Tank made the analogy between a picture that includes all discrete spikes and their arrival times as the ``quantum'' picture, with the rate-based treatment as the continuous, ``classical'' counterpart. While classical, rate-coding models can explain much of observed neural activity in behaving animals, significant information is omitted, and such models fail to capture many of the benefits of the manner in which neurons leverage the time domain for efficient computation and communication. Here we provide a model that is intended to play this role of a ``classical'' model of SOENs in the sense that discrete spikes are not explicitly treated. We find that while the model derived  is capable of rate coding, it is not limited to rate coding. Numerical experiments indicate that it is possible to treat the entire system without reference to spikes in a manner that remains capable of representing rapid (although still continuous), pulsatile activity. The model can accurately capture a dendrite or neuron's response to a single synapse event as well as to coincidences and sequences between two events. The simple governing equations facilitate formal analysis and provide opportunities for algorithm development for artificial intelligence, while the ability to represent abrupt activity allows numerical simulations to remain closely connected to actual hardware performance. We do not take a novel position on the rate-versus-spike debate \cite{brette2015philosophy}; our bias toward the value of spike timing is implicit in the numerical examples chosen, and our appreciation of the value of a rate-based formalism is clear from our choice to seek such a model in the first place.

In Sec.\,\ref{sec:model} of this work, we derive the phenomenological model of dendrites from the circuit Lagrangian. Through inspection of the properties of the dendritic model, we motivate and perform the extension to the full phenomenological model of neurons communicating to dendrites via synapses and show that the form of the equations modeling neurons in this way is identical to that of the dendrite, thereby unifying the treatment of all components of the system in a common framework. We assess the accuracy relative to the model that includes spikes in Sec.\,\ref{sec:quantification}. We compare the formalism to that of classical neurodynamics in Sec.\,\ref{sec:classical_neurodynamics}, and we discuss several implications in Sec.\,\ref{sec:discussion}.

\section{\label{sec:model}The Phenomenological model}
We begin by summarizing the model of Ref.\,\onlinecite{shainline2023phenomenological} to provide a framework for its extension to spiking neurons. Reference \onlinecite{shainline2023phenomenological} motivated the model with a rate-equation picture, while here we begin from a Lagrangian formalism, as this provides a more rigorous foundation as well as a pathway to extension to energy models and treatment of large systems with the tools of statistical mechanics.

\subsection{\label{sec:derivation}Derivation of the Phenomenological Dendrite Model}
\begin{figure}[tbh]
\includegraphics[width=8.6cm]{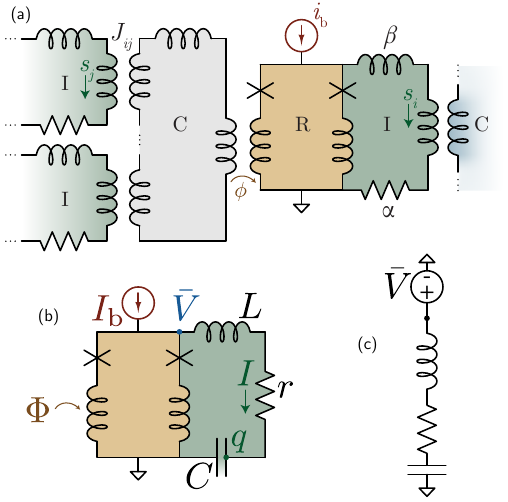}
\caption{\label{fig:circuits}Circuits considered in the phenomenological model. (a) Two dendritic integration loops (I) connected to a collection coil (labeled C), which couples flux ($\phi$) into the receiving loop of another dendrite (R). The dendrite itself is the combination of the R and I loops. (b) The dendritic SQUID and output $LRC$ loop with circuit parameters labeled. The objective of the phenomenological model is to represent the activity of this circuit without reference to the rapid JJ dynamics. (c) The circuit considered in the model, where the activity of the SQUID has been abstracted into the voltage source $\bar{V}$.}
\end{figure}
The circuit instantiation of a dendrite in the SOENs platform is shown in Fig.\,\ref{fig:circuits}. In part (a) of the figure, the integration loops (labeled I) from two input dendrites are shown feeding into the active dendrite under consideration through a collection coil (labeled C). The signals in the loops ($s$) and the coupled flux between dendrites ($\phi$) are labeled in terms of dimensionless quantities that will be derived shortly. Part (b) of Fig.\,\ref{fig:circuits} shows the active core of the dendrite with circuit parameters labeled using conventional circuit quantities, which will be used in the derivation. The objective of the phenomenological model is to capture the essential behaviors of this circuit without resorting to the full solution of the Josephson equations that describe the superconducting wave function on the picosecond timescale. The full circuit equations are numerically intensive and provide little insight into the computations of the dendrite. We would like to abstract the circuit to an input-output device where the output, $I$, can be efficiently computed in terms of the inputs, $\Phi$ and $I_\mathrm{b}$. To this end, we consider just the right half of the circuit, the $LRC$ branch, which we refer to as the integration loop. The equivalent circuit is shown in Fig.\,\ref{fig:circuits}(c).

We can write down the Lagrangian of the $LRC$ branch as \cite{wells1938application, goldstein2002classical} $\mathcal{L} = \mathcal{T}-\mathcal{V}$ with $\mathcal{T} = L\dot{q}^2/2$ and $\mathcal{V} = q^2/2C$, where $q$ is the charge on the capacitor, and $\dot{q}$ is its time derivative ($I=\dot{q}$ is the current). The resistor introduces a damping term, $\mathcal{F} = r\dot{q}^2/2$ \cite{goldstein2002classical}. The inductance, $L$, capacitance, $C$, and resistance, $r$, are assumed constant. The central concept of the phenomenological model is to treat the left half of the circuit, the SQUID, as an external electromotive force, $\mathcal{E}$. Once the proper form of this electromotive force is found, the equation of motion can be obtained from the Euler-Lagrange equation \cite{goldstein2002classical}:
\begin{equation}
\label{eq:euler-lagrange}
\frac{d}{dt}\left( \frac{\partial \mathcal{L}}{\partial \dot{q}} \right) - \frac{\partial \mathcal{L}}{\partial q} = \mathcal{E} - \frac{\partial \mathcal{F}}{\partial \dot{q}}.
\end{equation} 

To motivate the form for $\mathcal{E}$, we begin with the Josephson equation relating the voltage across a JJ to the rate of change of the phase of the superconducting wavefunction:
\begin{equation}
\label{eq:josephson_equation}
V(t) = \frac{\Phi_0}{2\pi}\,\frac{d\varphi}{dt},
\end{equation}
where $\Phi_0 = h/2e$ is the magnetic flux quantum. Integrating this equation gives
\begin{equation}
\label{eq:josephson_equation__integrated}
\int_0^{t_\mathrm{fq}} V(t)\,dt = \frac{\Phi_0}{2\pi}\,\int_0^{2\pi}d\varphi = \Phi_0.
\end{equation}
The limits of the integrals indicate that the phase of the superconducting wavefunction advances by $2\pi$ in a time that we define as $t_\mathrm{fq}$, which is the time required to produce a single flux quantum for a given temporal form of $V(t)$. In general, the voltage across each JJ in the SQUID will be a complicated function of the applied flux, the bias current, and the phase of the wavefunction across the other JJ. For the cases of neuromorphic interest considered here, the output of the SQUID will be coupled to an $LRC$ branch that will perform a low-pass filter on these high-speed dynamics. We thus take the crucial step that defines the phenomenological model, which is to replace $V(t)$ in Eq.\,\ref{eq:josephson_equation__integrated} with $\bar{V}(\bar{t})$, which we take to be constant over the short duration $t_\mathrm{fq}$. The quantity $\bar{V}$ does vary in time, but over timescales much longer than $t_\mathrm{fq}$. We denote the timescale over which $\bar{V}$ varies as $\bar{t}$. With this simplification, Eq.\,\ref{eq:josephson_equation__integrated} reduces to 
\begin{equation}
\label{eq:V_bar}
\bar{V}(\bar{t}) = \Phi_0\,G_\mathrm{fq}(\bar{t}),
\end{equation}
where $G_\mathrm{fq}(\bar{t}) = 1/t_\mathrm{fq}(\bar{t})$ is the rate of flux-quantum production defined on timescales long relative to $t_\mathrm{fq}$. The $\bar{t}$ notation serves to remind us that the model can only be expected to provide an accurate representation of the circuit over timescales long compared to flux-quantum production. It cannot be relied upon for picosecond dynamics, but it may be accurate on timescales of hundreds of picoseconds or longer, provided the $LRC$ branch performs as a low-pass filter, which is exactly the type of integrating behavior we desire from a dendrite. The important function is then $G_\mathrm{fq}(\bar{t})$, which we will discuss shortly.

To complete the phenomenological model, we make the identification $\mathcal{E} = \bar{V} = \Phi_0\,G_\mathrm{fq}$, and from Eq.\,\ref{eq:euler-lagrange} we obtain
\begin{equation}
\label{eq:equation_of_motion__dimensions}
\frac{d^2q}{dt^2} + \frac{r}{L}\,\frac{dq}{dt} + \frac{1}{LC}\,q = \frac{\Phi_0}{L}\,G_\mathrm{fq}.
\end{equation}
We define the characteristic frequency of the Josephson junctions as $\omega_c = 2\pi r_\mathrm{jj} I_c/\Phi_0$, where $r_\mathrm{jj}$ is the shunt resistance of a JJ in the resistively and capacitively shunted junction model \cite{van1998principles,kadin1999introduction,tinkham2004introduction,clarke2006squid}, and $I_c$ is the critical current of a single JJ. In terms of these quantities we can define a dimensionless charge, $\xi = (\omega_c/I_c)\,q$, and a dimensionless time, $t' = \omega_c\,t$. Moving to $t'$ gives $\frac{d^n\xi}{dt^n} = \omega_c^n\,\frac{d^n\xi}{dt'^n}$, so that Eq.\,\ref{eq:equation_of_motion__dimensions} becomes
\begin{equation}
\label{eq:equation_of_motion__dimensionless}
\beta\,\frac{d^2\xi}{dt'^2} + \alpha\,\frac{d\xi}{dt'} + \beta\left(\frac{\omega_{LC}}{\omega_c}\right)^2\,\xi = g_\mathrm{fq},
\end{equation}
where we have defined the dimensionless quantities $\beta = 2\pi L I_c/\Phi_0$, $\alpha = r/r_\mathrm{jj}$, the angular frequency of the $LC$ oscillator is $\omega_{LC} = (L C)^{-1/2}$, and $g_\mathrm{fq} = G_\mathrm{fq}/(\omega_c/2\pi)$. Equation \ref{eq:equation_of_motion__dimensionless} is the governing equation for a general SOEN dendrite, and it describes a damped, driven harmonic oscillator. The oscillatory component is useful for achieving resonances, subthreshold oscillations, and population synchronization. However, many dendrites will not include a capacitor and will be described instead in terms of a first order equation. In this case, $C\rightarrow\infty$ and $\omega_{LC} \rightarrow 0$. Introducing the dimensionless current $s = d\xi/dt' = I/I_c$ and the dimensionless flux $\phi = \Phi/\Phi_0$, we can write the phenomenological model for a first-order dendrite in the preferred form:
\begin{equation}
\label{eq:equation_of_motion__dimensionless__first-order}
\frac{ds}{dt} = \gamma\,g_\mathrm{d}(\phi,s; i_\mathrm{b}) - \frac{s}{\tau}.
\end{equation}
Here we dropped the prime on the dimensionless time variable as it is the only temporal variable in the remainder of the study. Equation \ref{eq:equation_of_motion__dimensionless__first-order} introduced alternative dimensionless quantities $\gamma = 1/\beta$, and $\tau = \beta/\alpha$ for convenience. The replacement $g_\mathrm{fq} \rightarrow g_\mathrm{d}$ was made to indicate that this source function applies to dendrites (as opposed to neurons). The reason for this replacement will be evident in Sec.\,\ref{sec:extension_to_neurons}. The source function depends on the dimensionless applied flux, $\phi$, and the dimensionless current, $s$, because these two quantities both affect the instantaneous electromotive force experienced by the circuit. $g_\mathrm{d}$ also depends on the dimensionless bias current, $i_\mathrm{b}$, as a parameter, as discussed in Ref.\,\onlinecite{shainline2023phenomenological}. 

\begin{figure}[tbh]
\includegraphics[width=8.6cm]{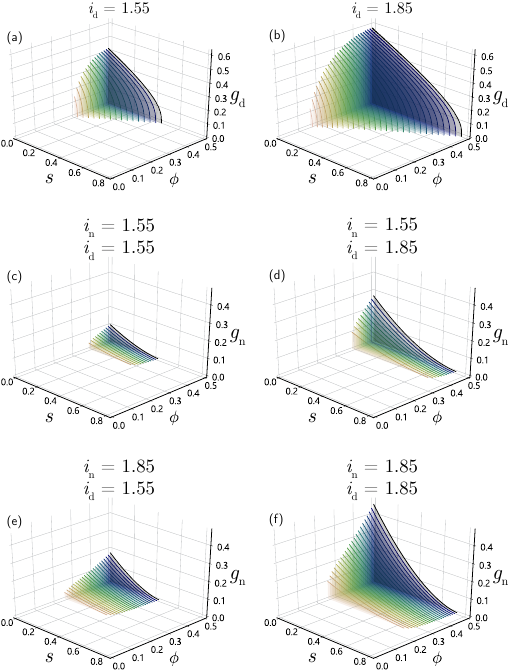}
\caption{\label{fig:source_functions}Source functions for direct connection from soma to dendrite. (a,b) Dendrite source functions for two different bias currents. (c-f) Neuron source functions for two different neuronal and two different dendritic bias currents.}
\end{figure}

To fully specify the model, we must identify the form of $g_\mathrm{d}(\phi,s)$. This can be accomplished with a series of numerical calculations, as described in Ref.\,\onlinecite{shainline2023phenomenological}. In brief, the full JJ circuit equations are solved for many values of $\phi$ and $i_\mathrm{b}$, and the rate of flux-quantum production is monitored as signal $s$ accumulates in the integration loop. A table of $g_\mathrm{d}$ is assembled as a function of $\phi$ and $s$ for multiple values of $i_\mathrm{b}$, which were referred to as rate arrays in Ref.\,\onlinecite{shainline2023phenomenological}. Here we change the terminology to refer to the same objects as source functions to avoid confusing the model with rate coding. In numerical implementation, the source function can be used as a look-up table or its form can be fit. Examples of the dendrite source functions are shown in Fig.\,\ref{fig:source_functions}(a) and (b).

The final aspect of the model relates to interactions. Dendrites are coupled to each other through flux. The coupling flux from dendrites indexed by $j$ to dendrite $i$ is given by
\begin{equation}
\label{eq:coupling}
\phi_i = \sum_{j=1}^n\,J_{ij}\,s_j,
\end{equation}
where $J_{ij}$ is a static coupling term determined by the mutual inductance that includes contributions from all the transformers present on the collection coil in Fig.\,\ref{fig:circuits}(a). The explicit form of $J_{ij}$ is given in Ref.\,\onlinecite{shainline2023phenomenological}. Equation \ref{eq:coupling} shows that coupling between dendrites is due to the signal in the integration loop of one dendrite being communicated as flux into the receiving loop of a subsequent dendrite. The signal in the subsequent dendrite is then obtained through the evolution of Eq.\,\ref{eq:equation_of_motion__dimensionless__first-order} with the signal from the first dendrite providing the flux $\phi$ and the signal from the second dendrite providing the $s$ term in the function $g_\mathrm{d}(\phi,s)$.

\begin{figure}[h!]
\includegraphics[width=8.6cm]{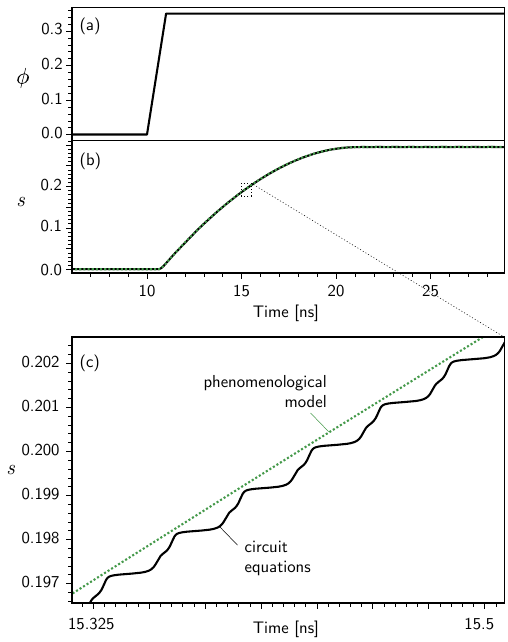}
\caption{\label{fig:ode_soen_comparison__first_phenomenological}Comparison of circuit response as calculated by first-principles circuit equations and phenomenological model. (a) The flux applied to the dendritic receiving loop. (b) The signal in the dendritic integration loop as calculated with the circuit equations (black, solid trace) and the first phenomenological model (green, dashed trace). (c) Zoom of a small region from (b) showing the the difference between the two models. For these calculations, $\beta_\mathrm{d}/2\pi = 10^3$, $\tau_\mathrm{d} = 250$\,ns, $\chi^2 = 2.5\times 10^{-6}$, $i_\mathrm{d} = 1.7$.}
\end{figure}

Given the approximation made in this model (replacing the true JJ voltages with $\bar{V}$ represented by the source function), it is not immediately clear under which circumstances the model will be useful or how accurate the result will be. Reference \onlinecite{shainline2023phenomenological} analyzed numerous cases and found accuracy of $10^{-4}$ as calculated with a distance metric ($\chi^2$) relative to the full circuit equations while achieving an increased speed of simulation of $10^4$. An illustrative example is shown in Fig.\,\ref{fig:ode_soen_comparison__first_phenomenological}. A step-function input is delivered to a dendrite, as seen in Fig.\,\ref{fig:ode_soen_comparison__first_phenomenological}(a). The accumulated signal in the dendritic integration loop is shown in Fig.\,\ref{fig:ode_soen_comparison__first_phenomenological}(b), with the black, solid trace showing the result of the full JJ circuit equations, and the green, dotted line showing the phenomenological model of Eq.\,\ref{eq:equation_of_motion__dimensionless__first-order}. The zoom in Fig.\,\ref{fig:ode_soen_comparison__first_phenomenological}(c) reveals the difference. The circuit equations give minute ripples, owing to the dynamics of the phase of the superconducting wave function across the two JJs, while the phenomenological model approximates the result with a smooth response. The discrepancy is tolerable in the present context because we care about the low-pass-filtered, integrated signal, which is accurate on the time scales of interest.

\subsection{\label{sec:extension_to_neurons}Extension of the Phenomenological Model to Neurons}
The phenomenological model described to this point has proven useful for simulating systems of superconducting optoelectronic neurons. However, each neuron's soma behaves in a manner that is not captured in this model. A soma is essentially a dendrite, with two important differences. First, the integration loop of a soma has a threshold, and when this threshold is reached, the integrated current is evacuated, and a refractory dendrite temporarily provides inhibitory flux to the receiving loop. Second, upon reaching threshold, an amplifier chain is activated that produces a pulse of light from a semiconductor transmitter circuit. The photons in this pulse then drive synaptic single-photon detectors to generate flux as input to downstream dendrites. These extensions of the dendrite circuit result in pulsatile, spiking behavior from the neuron. In the original model, all of these behaviors were modeled with circuit equations treating the transmitter circuit transistors and light source that provided faithful representation of the full spike response of the soma and the receiving synapses.

\begin{figure}[tbh]
\includegraphics[width=8.6cm]{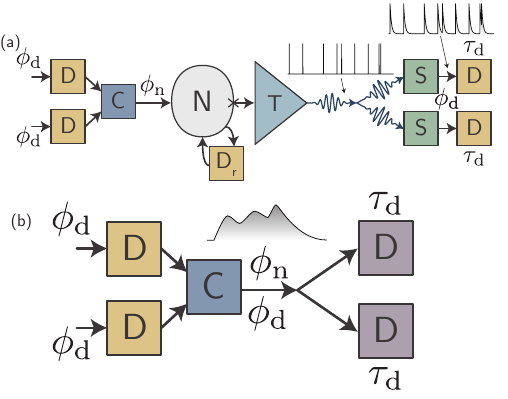}
\caption{\label{fig:schematic}Schematic of the phenomenological extension. (a) The original phenomenological model, with dendrites ($\mathsf{D}$) coupled into a collection coil ($\mathsf{C}$), which feeds into the neuron cell body ($\mathsf{N}$). The input flux to dendrites is $\phi_\mathrm{d}$, and that to the neuron is $\phi_\mathrm{n}$. Upon reaching somatic threshold, the neuron activates the refractory dendrite ($\mathsf{D_r}$), which delivers inhibitory feedback to the soma to temporarily increase the threshold for firing, and the transmitter ($\mathsf{T}$) is activated, which produces a photonic action potential that delivers light to downstream synapses ($\mathsf{S}$) that are coupled to output dendrites with time constant $\tau_\mathrm{d}$ that can vary across dendrites. In response to sustained input flux, the neuron will produce a train of action potentials, as illustrated by the stream of spikes between the transmitter and downstream synapses. (b) The full phenomenological model. In contrast to (a), the output flux from the collection coil, $\phi_\mathrm{n}$, is given directly to downstream dendrites, and these dendrites produce signals based on this flux using a different source function, and hence these boxes are colored red instead of yellow as used for dendrites not receiving flux directly from neurons. For these dendrites, $\phi_\mathrm{d} = \phi_\mathrm{n}$.}
\end{figure}
Here we develop an alternative approach to modeling neurons that avoids treating spikes in much the same way we avoided treating JJ dynamics in the first phenomenological model. The concept is depicted in Fig.\,\ref{fig:schematic}. Figure \ref{fig:schematic}(a) represents a neuron with two dendrites feeding flux into a passive, linear collection coil, which couples into the soma. The schematic illustrates that the transmitter produces spike events that deliver photons to downstream synapses coupled to dendrites. By contrast, in Fig.\,\ref{fig:schematic}(b) the circuit is the same up to the collection coil, but the entire soma, transmitter, and synapses are omitted. We seek a model in which the signal generated in downstream dendrites can be obtained exclusively through consideration of the applied flux to the neuronal receiving loop, $\phi_\mathrm{n}$. Importantly, this flux is an analog, continuous function of time that is not erased upon neuronal firing, just as the flux applied to any dendrite provides a persistent input drive, independent of the output signal, $s$. This analog signal is indicated in Fig.\,\ref{fig:schematic}(b) where the flux input to the neuronal receiving loop, $\phi_\mathrm{n}$ becomes also the flux to each of the downstream dendritic receiving loops, $\phi_\mathrm{d}$. This signal is shown schematically as a continuously varying signal in place of a train of discrete spikes. This concept, coupled with the image of the underlying approximation captured in Fig.\,\ref{fig:ode_soen_comparison__first_phenomenological}(c) motivates us to seek a phenomenological model for the soma, transmitter, and synapses in a nearly identical fashion to the model constructed for the dendrites, except targeting yet a slower time scale. While the first phenomenological model accepts ignorance of dynamics faster than 100\,ps by ignoring Josephson behavior on the 1\,ps to 10\,ps timescale, here we accept ignorance of dynamics faster than 50\,ns by ignoring the precise synaptic response on the 1\,ns to 10\,ns timescale. 

To construct such a model, we repeat the steps described in Sec.\,\ref{sec:derivation}. We again seek a differential equation describing the rate of change of the signal in a dendrite, but instead of taking the driving term to be a function of the flux applied to that dendrite, we let the flux to the neuronal receiving loop of the upstream soma be the input driving argument. We conjecture that this will be possible with an equation of the form
\begin{equation}
\label{eq:equation_of_motion__dimensionless__first-order__neuron}
\frac{ds}{dt} = \gamma\,g_\mathrm{n}(\phi,s) - \frac{s}{\tau}.
\end{equation}
We now must find the new $g_\mathrm{n}$, which is related to but distinct from $g_\mathrm{d}$. We determine $g_\mathrm{n}$ in exactly the same way as we did for the dendrite: we apply a constant flux to the neuronal receiving loop of a soma, and we observe the rate of flux-quantum production in a downstream dendrite that is coupled to the neuron via a transmitter and synaptic receiver. We use the first phenomenological model of Ref.\,\onlinecite{shainline2023phenomenological}, complete with transmitter and synaptic dynamics, to construct this source function. 

Due to the added complexity of the full neuronal circuit resulting from the circuit values describing the soma, the resulting source function is specified by additional parameters, including the value of the somatic threshold, $s_\mathrm{th}$, the bias current to the soma, $i_\mathrm{n}$, properties of the refractory dendrite, and properties of the synaptic receiver. Most of these parameters can be fixed at reasonable values. For the present study, we consider only $s_\mathrm{th} = 0.2$, one fixed refractory dendrite design, and a default synapse configuration. The source function takes as arguments $\phi_\mathrm{n}$, $s$, and the slowly varying or constant parameters $i_\mathrm{n}$ and $i_\mathrm{d}$. Examples of these neuronal source functions are shown in Fig.\,\ref{fig:source_functions}(c)-(f) for two values of $i_\mathrm{n}$ and two values of $i_\mathrm{d}$. Increasing the bias to the neuron shifts the flux threshold to lower values, while increasing the bias to the receiving dendrite increases the signal saturation value. These behaviors are slightly different from the source function for dendrites, where increasing the bias in that case reduces the flux threshold and also increases the signal saturation value. While the specific form of the source function $g$ depends on whether the dendrite in question is receiving flux from other dendrites ($g_\mathrm{d}$) or from a neuron ($g_\mathrm{n}$), the form of the equation is identical, which we write in general as
\begin{equation}
\label{eq:main_equation}
\frac{ds}{dt} = \gamma\,g(\phi,s) - \frac{s}{\tau}.
\end{equation}
The specific values of $\gamma$ and $\tau$ can be chosen independently for each dendrite across a wide range of values \cite{khan2022superconducting}.

\begin{figure}[t!]
\includegraphics[width=8.6cm]{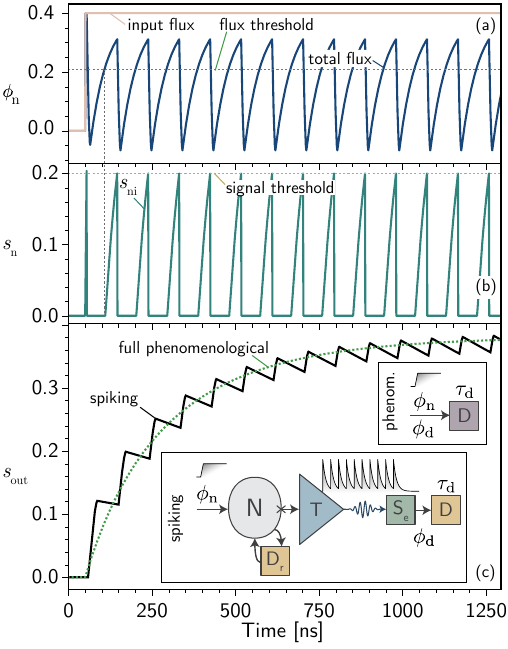}
\caption{\label{fig:step_response}Second phenomenological model concept illustrated with the response to an input flux step function. (a) The soma dynamics as calculated with the model of Ref.\,\onlinecite{shainline2023phenomenological}. The externally applied flux input to the soma is a step function above the somatic flux threshold. The total flux is the external input plus the feedback flux from the refractory dendrite. (b) The signal generated in the neuronal integration loop due to the flux drive. The saw-tooth behavior results from purging of signal upon reaching threshold, followed by refraction and reaccumulation. (c) Comparison of the responses of downstream recipient dendrites calculated with the spiking and phenomenological models. $s_\mathrm{out}$ is the signal integrated in the output dendrite. The insets show the neural circuit schematics for the two cases. The spiking model requires calculation of the soma traces shown in (a) and (b) as well as the response of the synaptic single-photon detector, while the phenomenological model only uses the input flux signal, with the dynamics of the neuronal integration loop, refractory dendrite, and synaptic detector encoded in $g_\mathrm{n}$. The $\chi^2$ between the signals in the output dendrites calculated with the spiking and phenomenological models was $1.91\times 10^{-3}$. $\beta_\mathrm{in}/2\pi = 10^3$, $\tau_\mathrm{in} = 250$\,ns, $\beta_\mathrm{out}/2\pi = 10^4$, $\tau_\mathrm{out} = 1.25$\,\textmu s. For the input dendrite, soma, and output dendrite, $i_\mathrm{b} = 1.7$.}
\end{figure}

The result is exemplified in Fig.\,\ref{fig:step_response}, which shows a soma being driven by a step function of input flux. The partial phenomenological model of Ref.\,\onlinecite{shainline2023phenomenological} is used to calculate the complete soma dynamics, shown in Fig.\,\ref{fig:step_response}(a) and (b). The output signal in a downstream dendrite is shown in Fig.\,\ref{fig:step_response}(c) for both the partial and full phenomenological models. Schematics for the spiking and phenomenological neural circuits used to make this figure are shown in the insets. In the spiking case, the step-function applied flux to the neuronal receiving loop, $\phi_\mathrm{n}$, is input to the soma, which produces a train of spikes at a well-defined rate based on the dynamics of the soma itself, the refractory dendrite, and the transmitter circuit. In the full phenomenological case, this step-function input is applied directly to the downstream dendrite so that $\phi_\mathrm{n} = \phi_\mathrm{d}$, and the only difference between a dendrite receiving spikes from a neuron and a dendrite receiving flux from another dendrite is the specific form of the source term, $g_\mathrm{d}$ versus $g_\mathrm{n}$.

For the spiking case, circuit operation can be understood as follows. If the applied flux exceeds the flux threshold of the somatic dendrite [Fig.\,\ref{fig:step_response}(a)], signal will begin to accumulate in the neuronal integration loop [Fig.\,\ref{fig:step_response}(b)], and if that signal reaches the current threshold of the integration loop, the neuron will produce spikes that deliver impulses to downstream synapses, and the neuron will also activate the refractory dendrite. The sum of the external input flux and the refractory inhibitory flux can be seen in Fig.\,\ref{fig:step_response}(a), labeled ``total flux''. The saw-tooth behavior of the neuronal integration loop seen in Figs.\,\ref{fig:step_response}(a) and (b) results from the steady drive paired with reset dynamics upon reaching threshold. It is precisely these fast oscillations that we avoid modeling with the full phenomenological model, analogous to the treatment in which we glossed over the fast JJ dynamics in a dendrite. The results of the two approaches are compared in Fig.\,\ref{fig:step_response}(c), which is strikingly similar to Fig.\,\ref{fig:ode_soen_comparison__first_phenomenological}(c) except that the time scale is slower by nearly four orders of magnitude. We are applying the phenomenological concept, based on low-pass filtered outputs, hierarchically across scales of the network, first to individual dendrites, and then between neurons. It may eventually prove necessary to employ the same trick at the level of populations.

While this model does give up some accuracy in tracking the fast, saw-tooth response of dendrites, it provides a helpful unification across the system: only dendrites must be considered; neither synapses nor neurons are explicitly modeled; plasticity functions can be treated using dendrites that obey the exact same equation; and many benefits of spiking neurons are retained, while the complications of spike discontinuities are not present in the mathematics. Before exploring various cases to test the accuracy and limits of the model, let us further describe why this unification is conceptually and numerically helpful.

\subsection{\label{sec:modeling_systems}Modeling Systems with the Phenomenological Framework}
Consider a neural system implemented in superconducting optoelectronic hardware. The network comprises a set of $N$ dendrites, with the state of dendrite $i$ given by $s_i(t)$, which obeys Eq.\,\ref{eq:main_equation} with the appropriate choice of $g_\mathrm{d}$ or $g_\mathrm{n}$ depending on whether the dendrite is a common dendrite in a neuron's arbor or a synaptic dendrite that receives signal directly based on the flux applied to the upstream soma. The flux applied to dendrite $i$ that generates signal is calculated from
\begin{equation}
\label{eq:main_equation__coupling}
\phi_i = \sum_j J_{ij}\,s_j\,+\,\phi_i^\mathrm{ext},
\end{equation}
where $J_{ij}$ is an element of the static coupling matrix established by a transformer that can be learned in design but is fixed upon fabrication, and $\phi_i^\mathrm{ext}$ is the external drive that excites dendrite $i$. Dendrites for which $\phi_i^\mathrm{ext}\ne 0$ are referred to as input dendrites. We can place all $s_i$ in a column vector, $\mathbf{s}(t) \equiv \left( s_1(t)\,\cdots\,s_N(t) \right)^T$, which is the state vector of the system. $\mathbf{s}(t)$ describes the system as a point in configuration space, and its time evolution traces a trajectory. We place all $J_{ij}$ in a square coupling matrix, $\mathbf{J}$. The flux vector is given by
\begin{equation}
\label{eq:flux_vector}
\boldsymbol{\phi}(t) = \mathbf{J}\,\mathbf{s}(t)\,+\,\boldsymbol{\phi}_\mathrm{ext}(t).
\end{equation}
We can similarly organize the elements $\gamma_i$ and $\tau_i$ in column vectors $\boldsymbol{\gamma}$ and $\boldsymbol{\tau}$, and write the equation of flow through configuration space as
\begin{equation}
\label{eq:main_equation__vector_form}
\frac{d\mathbf{s}}{dt} = \boldsymbol{\gamma}\odot\mathbf{g}(\boldsymbol{\phi},\mathbf{s}) - \frac{\mathbf{s}}{\boldsymbol{\tau}},
\end{equation}
where $\mathbf{g}(\boldsymbol{\phi},\mathbf{s})$ is a vector function that takes two vector inputs and returns a vector output computed component-wise. The ``$\odot$'' notation refers to component-wise multiplication (Hadamard product), and $\mathbf{s}/\boldsymbol{\tau}$ is component-wise division. 

With this unified model and notation, the numerical algorithm becomes simple, transparent, and highly parallelizable. The initial condition, $\mathbf{s}(t_0)$, must be provided, and $\mathbf{s}(t_0) = 0$ is common. The network is inactive prior to the introduction of driving flux. The state of the network at subsequent time steps indexed by $t_p$ is obtained as follows:
\begin{enumerate}
\item With known $\boldsymbol{\phi}_\mathrm{ext}$ and $\mathbf{J}$, calculate $\boldsymbol{\phi}(t_{p+1}) = \mathbf{J}\,\mathbf{s}(t_p)\,+\,\boldsymbol{\phi}_\mathrm{ext}(t_{p+1})$. This operation is just the matrix-vector multiplication of $\mathbf{J}\,\mathbf{s}$, where $\mathbf{J}$ is an extremely sparse matrix. 
\item Call a function $\mathbf{g}(t_{p+1}) = \mathbf{g}[\boldsymbol{\phi}(t_{p+1}),\mathbf{s}(t_p)]$ that computes the signal to be added to all dendrites at the next time step. Each of these is independent, so the operation is highly parallelizable.
\item Obtain $\mathbf{s}(t_{p+1})$ using a discrete form of Eq.\,\ref{eq:main_equation__vector_form} and the values of $\mathbf{g}(t_{p+1})$, $\mathbf{s}(t_p)$, $\boldsymbol{\gamma}$, and $\boldsymbol{\tau}$.
\item Return to first step and repeat forward Euler through the time grid.
\end{enumerate}

This algorithm is efficient for several reasons. Step 1 involves matrix-vector multiplication, $\mathbf{J}\,\mathbf{s}$, but $\mathbf{J}$ is extremely sparse, so fast algorithms that leverage graphics or tensor processing units can be used \cite{alahmadi2020performance}. Step 2 requires calculation of $\mathbf{g}[\boldsymbol{\phi}(t_{p+1}),\mathbf{s}(t_p)]$, and this can be distributed over as many cores as one can access, because the computation for each element is independent from the other elements at each time. Step 3 is simply arithmetic---element-wise multiplication and subtraction---which again can be easily parallelized. In this form, the algorithm reduces to one sparse matrix-vector multiplication and a few parallelizable operations. The entire algorithm is well-matched to graphics or tensor processing units. In addition to being faster than the spiking algorithm, which requires special treatment on steps in which an action potential is produced, it is also significantly more transparent, with no thresholds, if statements, synaptic response functions, or transmitter models. The model only references dendrites that all evolve with the same differential equation. The entire state of the network at time $t$ is specified by $\mathbf{s}(t)$, which is a list of real-valued numbers between zero and one. There is no need to track discontinuous quantities or tabulate spike times. We next assess the accuracy of the model in several cases of relevance to neural systems.

\section{\label{sec:quantification}Quantifying Accuracy and Domain of Applicability}
Based on the premise of the model, we expect it to be accurate only for dendrites that integrate over time scales relatively long compared to the inter-spike interval. Nevertheless, it is worth assessing the accuracy of the model when attempting to capture activity ranging from a single synapse event up to long trains of pulses. We begin by considering a monosynaptic neuron under various drive conditions and then consider a disynaptic neuron to assess the ability of the model to capture coincidences and sequences. To assess the accuracy, we compare the full phenomenological model to the partial phenomenological model that includes spiking, which was compared to circuit equations in Ref.\,\onlinecite{shainline2023phenomenological}. We quantify accuracy with a $\chi^2$ of the form
\begin{equation}
\label{eq:chi_squared}
\chi^2 = \frac{ \sum_{p=1}^{n_t} \left| s_\mathrm{spike}(t_p) - s_\mathrm{phenom}(t_p) \right|^2 }{ \sum_{p=1}^{n_t} \left| s_\mathrm{spike}(t_p) \right|^2 },
\end{equation}
where $p$ indexes the discrete numerical time steps, $n_t$ is the number of time steps, $s_\mathrm{spike}$ is the calculated value of the signal in the output dendrite using the first phenomenological model that faithfully represents the spiking activity of the soma, transmitter, and receiving synapse, while $s_\mathrm{phenom}$ is the calculated signal in the output dendrite using the full phenomenological model that omits spiking and instead generates signal using a source function driven by $\phi_\mathrm{n}$. This $\chi^2$ is not defined when $s_\mathrm{spike} = 0$.

For all simulations investigated with the full phenomenological model in this work, the soma was specified by $i_\mathrm{b} = 1.7$, $\beta_\mathrm{ni}/2\pi = 10^3$, $\tau_\mathrm{ni} = 50$\,ns, and $s_\mathrm{th} = 0.2$. The refractory dendrite had the same values of $i_\mathrm{b}$, $\beta$, and $\tau$.

\subsection{\label{sec:monosynaptic_neuron}Monosynaptic Neuron}
\begin{figure}[tbh]
\includegraphics[width=8.6cm]{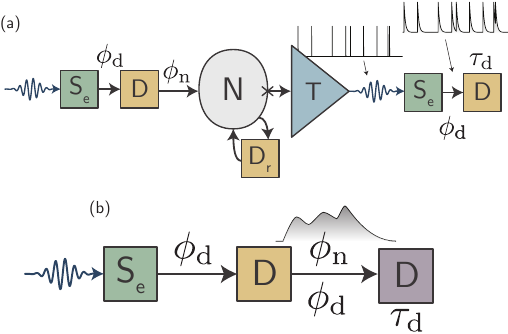}
\caption{\label{fig:schematic__monosynaptic}Schematics of the circuits under consideration for a monosynaptic neuron. (a) The first phenomenological model with a spiking soma. (b) The second phenomenological model with the input neural flux, $\phi_\mathrm{n}$, passed directly to the downstream dendrite.}
\end{figure}
The monosynaptic neuron under consideration is shown schematically in Fig.\,\ref{fig:schematic__monosynaptic}. The full spiking version is shown in Fig.\,\ref{fig:schematic__monosynaptic}(a), and the phenomenological version is shown in Fig.\,\ref{fig:schematic__monosynaptic}(b). 

\subsubsection{\label{sec:single_synapse_event}Single Synapse Event}
\begin{figure}[tbh]
\includegraphics[width=8.6cm]{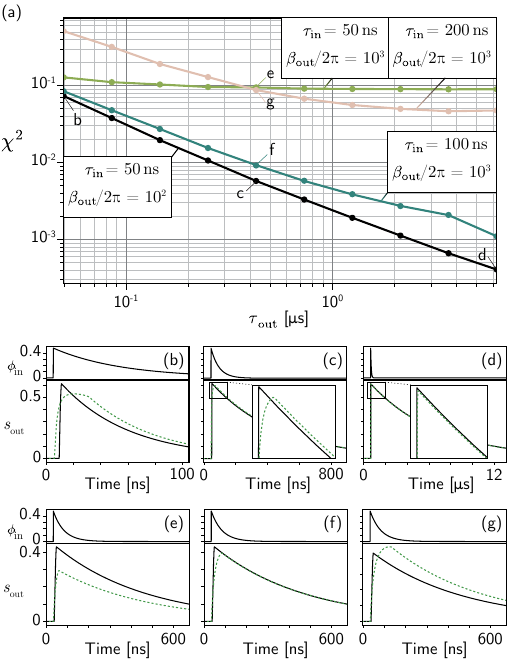}
\caption{\label{fig:sing_syn_event}Analysis of single input synapse events with $\beta_\mathrm{di}^\mathrm{in} = 2\pi\,10^2$ and $i_\mathrm{b}^\mathrm{n} = 1.7$ and $i_\mathrm{b}^\mathrm{d} = 1.75$. The time traces in (b)-(g) are from the data points inside the red boxes with the corresponding labels in (a).}
\end{figure}
To begin assessing the utility of the model, we compare the spiking version to the full phenomenological version when a single synapse event is input to a receiving dendrite on the neuron. For all spiking simulations in this section, the neuron circuit parameters are chosen such that the single input synapse event pushes the neuron above threshold, and the neuron produces one output spike, which is delivered to a downstream synapse coupled to a dendrite. The goal of the phenomenological model in this case is to accurately represent the output dendritic signal when all the model has access to is the time trace of input flux to the soma due to the input synapse event.

In biology, the temporal envelope of the post-synaptic signal is determined by the properties of neurotransmitter receptors (AMPA and NMDA for excitatory glutamate and GABA$_\mathrm{A}$ and GABA$_\mathrm{B}$ for inhibitory GABA), which have time constants in the range of 5\,ms to 100\,ms (shorter for AMPA and GABA$_\mathrm{A}$, longer for NMDA and GABA$_\mathrm{B}$). In SOENs, the analogous time constants are determined by the decay of the dendritic integration loop, which can be very fast (less than 1\,ns), or extended to biological time scales, with 5.8\,ms demonstrated in Ref.\,\onlinecite{khan2022superconducting}. The simulations in this section consider dendrites with a time constant in the range of 50\,ns to 6.25\,\textmu s responding to the arrival of a single action potential, intended to cover the primary range of circuit parameters used in common SOENs applications.

Figure \ref{fig:sing_syn_event}(a) shows $\chi^2$ calculated as a function of the time constant of the output dendrite, $\tau_\mathrm{out}$, for several values of $\tau_\mathrm{in}$, $\beta_\mathrm{out}$, and $i_\mathrm{d}$. The common trend is that the representation becomes more accurate as $\tau_\mathrm{out}$ increases, which is to be expected, as this model is intended to capture low-pass dynamics. Example time traces are shown corresponding to the data points labeled b-g in Fig.\,\ref{fig:sing_syn_event}(a). In the first time trace, Fig.\,\ref{fig:sing_syn_event}(b), both the input dendrite time constant and the output dendrite time constant are 50\,ns, which is close to the minimum interspike interval of SOENs. In this case we would expect the accuracy to be poor, as almost no low-pass filtering is occurring in the circuit, and indeed the value of $\chi^2$ is relatively low at 0.07. However, inspection of the temporal response reveals that while the detailed structure of the signal in the output dendrite does not accurately match that of the spiking model, the qualitative behavior is similar: an input synapse event to the neuron evokes a pulse of activity in the downstream dendrite. For some modeling purposes, this numerical error may be intolerable, and the full spiking model must be used. But for many modeling applications and algorithm development, such accuracy may be sufficient. Moreover, this limit of rapidly decaying signals on the downstream dendrite is not likely to be common in practice, as the primary function of most dendrites is to integrate signals over time, for which purpose time constants longer than the most rapid interspike interval are required. For example, Fig.\,\ref{fig:sing_syn_event}(c) shows the response when the decay time constant of the downstream dendrite is 427\,ns, which is about 10 times the minimum interspike interval. In this case the error is $5.7\times 10^{-3}$, and the qualitative representation of the resulting dendritic response is excellent. The representation further improves with $\tau_\mathrm{di}^\mathrm{out} = 6.25$\,\textmu s, as shown in Fig.\,\ref{fig:sing_syn_event}(d), with $\chi^2 = 3.9\times 10^{-4}$.

The data in Fig.\,\ref{fig:sing_syn_event}(b)-(c) are from a circuit in which the input dendrite had a low-capacity integration loop, with $\beta_\mathrm{in}/2\pi = 10^2$. The data in Fig.\,\ref{fig:sing_syn_event}(e)-(g) are from a circuit in which the input dendrite had a higher-capacity integration loop, with $\beta_\mathrm{in}/2\pi = 10^3$, so the loop can integrate the inputs from several synapse events before saturating. The time traces in Fig.\,\ref{fig:sing_syn_event}(e)-(g) are from scenarios in which the time constant of the dendrite providing input to the soma varies from 50\,ns to 200\,ns. The quantitative agreement for the 100\,ns case is much better than the other two, with the phenomenological model underestimating the response for the 50\,ns drive and overestimating the response for the 200\,ns drive. Nevertheless, the qualitative response is in accord in all three cases. 

\begin{figure}[t!]
\includegraphics[width=8.6cm]{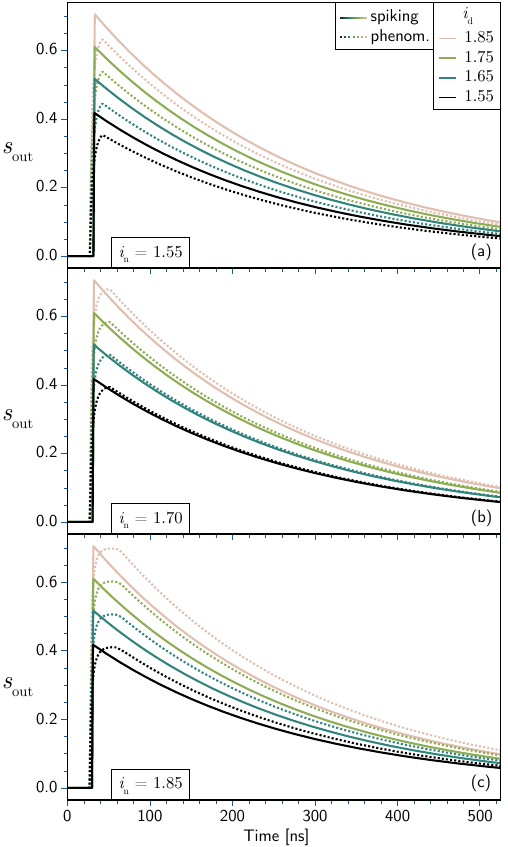}
\caption{\label{fig:sing_syn_event__vary_ib}Responses of single synapse events with differing neuron and dendrite bias currents. (a) The neuron bias $i_\mathrm{n} = 1.55$. (b) The neuron bias $i_\mathrm{n} = 1.7$. (c) The neuron bias $i_\mathrm{n} = 1.85$. For all cases shown here, $\beta_\mathrm{in}/2\pi = 10^2$, $\tau_\mathrm{in} = 50$\,ns, $\beta_\mathrm{out}/2\pi = 10^2$, and $\tau_\mathrm{out} = 250$\,ns. $i_\mathrm{in} = 1.7$.}
\end{figure}

It is useful to understand how accuracy may depend on bias current, as this circuit parameter may be used to realize different dendrite or neuron classes or as a plasticity mechanism for training and learning. Figure \ref{fig:sing_syn_event__vary_ib} compares the phenomenological and spiking models for various neuron and dendrite bias currents. Here we fix the input time constant to 50\,ns and the output time constant to 250\,ns. The output loop had $\beta_\mathrm{out}/2\pi = 10^2$, so it saturates quickly. Similarly to Fig.\,\ref{fig:sing_syn_event}(e)-(g), the phenomenological model may overshoot or undershoot the spiking model. Panels (a)-(c) of Fig.\,\ref{fig:sing_syn_event__vary_ib} each show four values of the bias current to the soma. For the spiking model, this has very little effect on the output in this specific instance of a single input synapse event; the latency of spike production is slightly impacted. However, this bias current does affect the phenomenological model, as it determines the subtleties of the source function [Fig.\,\ref{fig:source_functions}(c)-(f)]. It is not clear why the model is more accurate for some bias currents than others.

Because this model abstracts away spike activity, it may be tempting to consider it a rate-coded model. In a rate-coded model, the rate is typically defined either as the reciprocal of the inter-spike interval (which requires two or more spikes to be defined) or as the number of spikes in a moving-window average (which can be defined for a single spike). In the present case of Figs.\,\ref{fig:sing_syn_event} and \ref{fig:sing_syn_event__vary_ib} generated with Eq.\,\ref{eq:main_equation}, neither of these definitions of rate coding are applicable. Only one spike is present, so the first definition of a rate cannot apply, and no information about an average number of spikes in a moving time window is provided to the model. However, Eqs.\,\ref{eq:equation_of_motion__dimensionless__first-order} and \ref{eq:equation_of_motion__dimensionless__first-order__neuron} are unambiguously rate equations, so we must understand their proper interpretation. In the case of a dendrite, the phenomenological model given by Eq.\,\ref{eq:equation_of_motion__dimensionless__first-order} represents the rate of change of signal in the integration loop of that dendrite due to flux applied to the receiving loop of that dendrite as well as passive, exponential signal decay in the integration loop. In the case of a neuron, the phenomenological model given by Eq.\,\ref{eq:equation_of_motion__dimensionless__first-order__neuron} represents the rate of change of signal in the integration loop of a downstream dendrite due to flux applied to the receiving loop of the neuron's soma as well as passive, exponential signal decay in the downstream dendrite's integration loop. Information regarding spike rates is not provided to or generated by the model. The phenomenological model is not a model of neuronal rate coding, but rather a model of dendritic signal rates of change due to driving fluxes and passive leaks.

\subsubsection{\label{sec:pulse_trains}Pulse Trains}
\begin{figure}[tbh]
\includegraphics[width=8.6cm]{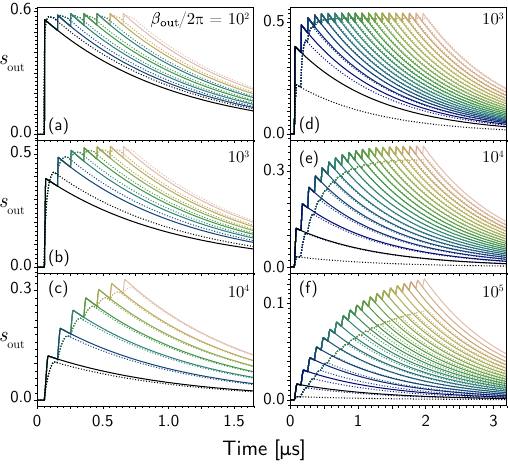}
\caption{\label{fig:cascades}Response of the phenomenological model to a periodic input burst. (a)-(c) The case with $\beta_\mathrm{in}/2\pi = 10^2$, $\tau_\mathrm{in} = 150$\,ns, and $\tau_\mathrm{out} = 1$\,\textmu s. (a) $\beta_\mathrm{out}/2\pi = 10^2$. (b) $\beta_\mathrm{out}/2\pi = 10^3$. (c) $\beta_\mathrm{out}/2\pi = 10^4$. (d)-(f). $\beta_\mathrm{in}/2\pi = 10^3$, $\tau_\mathrm{in} = 250$\,ns, $\tau_\mathrm{di}^\mathrm{out} = 1.25$\,\textmu s. (d) $\beta_\mathrm{out}/2\pi = 10^3$. (e) $\beta_\mathrm{out}/2\pi = 10^4$. (f) $\beta_\mathrm{out}/2\pi = 10^5$. In all cases, $i_\mathrm{b}^\mathrm{n} = i_\mathrm{b}^\mathrm{d} = 1.7$, solid lines are the spiking model, and dashed lines are the phenomenological model. The input interspike interval was 100ns.}
\end{figure}
Next consider a train of synapse events input to the monosynaptic neuron of Fig.\,\ref{fig:schematic__monosynaptic}. The response to a neuronal pulse train evoked by a constant flux input to the soma was shown in Fig.\,\ref{fig:step_response}. Here we consider a temporally extended input drive due to a train of input pulses delivered to a synapse coupled to a dendrite with output fed into the soma. In this scenario, the neuron is acting as a repeater, with each input synapse event evoking a spike, and we monitor the response of the output dendrite to those spikes induced in the neuron.

First consider the response to an input train of seven synapse events with uniform interspike interval, as shown in Fig.\,\ref{fig:cascades}. Here the responses to multiple input trains of different numbers of pulses are shown overlaid so one can see how the accuracy changes as the pulse train proceeds. In Fig.\,\ref{fig:cascades}(a)-(c) the input dendrite has a small loop capacity, with $\beta_\mathrm{in}/2\pi = 10^2$ and relatively short input time constant of $\tau_\mathrm{in} = 150$\,ns. Responses are shown for output loops of varying inductance parameter from $\beta_\mathrm{di}^\mathrm{out}/2\pi = 10^2$ to $\beta_\mathrm{di}^\mathrm{out}/2\pi = 10^4$. For the smaller output loops, the strength of the initial synapse events is slightly overestimated by the phenomenological model, while for the largest loop the initial effect is slightly underestimated, but in all cases the overall accuracy in response to the train becomes reasonably high as the train proceeds. 

Figure \ref{fig:cascades}(d)-(f) shows responses in the case of a larger input integration loop, $\beta_\mathrm{in}/2\pi = 10^3$, capable of integrating more synapse events, with $\tau_\mathrm{in} = 250$\,ns. Here a train of 20 input synapse events is considered. The responses are shown for output loops of varying inductance parameter from $\beta_\mathrm{out}/2\pi = 10^3$ to $\beta_\mathrm{out}/2\pi = 10^5$. While the response of the initial synapse events is undervalued in all cases, the signal catches up through the pulse train for the case of $\beta_\mathrm{out}/2\pi = 10^3$, while it remains slightly low for $\beta_\mathrm{out}/2\pi = 10^4$, and the error is even more pronounced for $\beta_\mathrm{di}^\mathrm{out}/2\pi = 10^5$. While a loop of $\beta_\mathrm{out}/2\pi = 10^5$ is so (spatially) large as to rarely be used in practice, it is important to be aware that in this region of circuit parameter space the phenomenological model does underestimate the response to the entire pulse train. Still, the qualitative agreement is likely sufficient for much theoretical analysis and algorithm development.

\begin{figure}[t!]
\includegraphics[width=8.6cm]{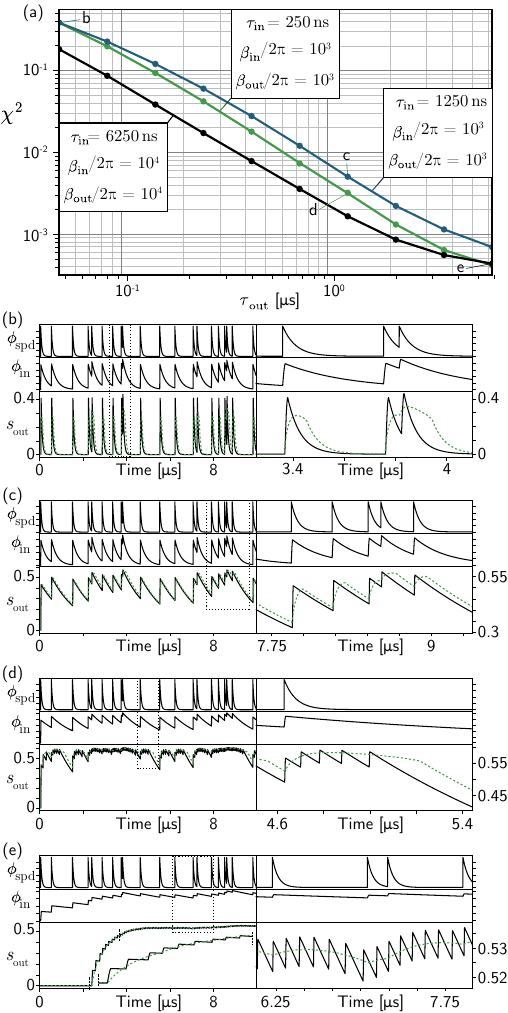}
\caption{\label{fig:train_syn_events}Train of synapse events. (a) $\chi^2$ versus $\tau_\mathrm{out}$. Points corresponding to the time traces in (b-e) are labeled. (b-e) Time traces with the temporal zoom from the region with the dotted lines shown at the right. In part (e), the initial rising section of the signal between the vertical dashed lines is shown as an inset.}
\end{figure}

Next consider a train of 20 aperiodic input synapse events with interspike intervals drawn randomly with uniform probability between 50\,ns and 1\,\textmu s. Figure \ref{fig:train_syn_events} shows similar data to Fig.\,\ref{fig:sing_syn_event} with $\chi^2$ plots accompanied by illustrative time traces. As in Fig.\,\ref{fig:sing_syn_event}, $\chi^2$ drops markedly for longer output time constant, consistent with the principle of the model as a low-pass filter. Figure \ref{fig:train_syn_events}(b) is the case with the worst accuracy, wherein the input dendrite had $\beta_\mathrm{in}/2\pi = \beta_\mathrm{out}/2\pi = 10^3$, and like Fig.\,\ref{fig:sing_syn_event}(b), the agreement is qualitatively reasonable but quantitatively poor. The phenomenological model tracks the true spike model as individual synapses perturb the state of the neuron, but the output signal is much more rounded than the sharp peaks of the spiking model. Figures \ref{fig:train_syn_events}(c) and (d) also show the case of $\beta_\mathrm{in}/2\pi = \beta_\mathrm{out}/2\pi = 10^3$. In (c) the input time constant was 250\,ns, while for (d) it was 1.25\,\textmu s, and for both cases the output time constant was 1.25\,\textmu s. Here the $\chi^2$ values are much better, well under 1\,\%. For the faster input circuit of (c), the phenomenological model accurately tracks the rising and falling response of the spiking model, with low-pass filtering evident in the temporal zoom. For the slower input circuit of (d), the phenomenological model closely matches the time averaged behavior, and in this case the model closely matches what would be expected from rate coding. The final specific instance shown in (e) is the case of long integration times ($\tau_\mathrm{in} = \tau_\mathrm{out} = 6.25$\,\textmu s) and high-capacity loops ($\beta_\mathrm{in}/2\pi = \beta_\mathrm{out}/2\pi = 10^4$). The accuracy is high at $\chi^2 = 4.4\times 10^{-4}$. Significant low-pass filtering occurs, and again we are in the rate-coding domain. Due to the large input loop capacity, the signal from multiple synapse events must be integrated before the input dendrite drives the soma above threshold and signal begins to accumulate in the output dendrite, just after 2\,\textmu s. Upon reaching threshold, both the spiking and phenomenological models begin accumulating signal at nearly identical times and comparable rates. The inset to Fig.\,\ref{fig:train_syn_events}(e) shows the rising signal in the initial integration region. After roughly 2\,\textmu s of integration, the output dendrite is saturated, and signal changes are minute. The average spiking activity is sufficient to keep the dendrite saturated despite the variability in interspike interval. The temporal zoom shows that the phenomenological model traces the average signal level of the spiking model, as we would expect from a rate-coding picture with instantaneously updated rate term, but the phenomenological model does not track the full saw-tooth response of the spiking model. From another perspective, one may argue the phenomenological model does an excellent job of representing the smoothly varying input to the soma, while the spiking model does a reasonable job approximating the smooth function, given the limitation that it can only send discrete events from the soma to downstream synapses.

Considering all examples presented here, we conclude that the phenomenological model is capable of but not limited to rate coding. Rate coding arises in the specific cases where the drive signal is extended in time and the output dendrite has a long time constant relative to the interspike interval. However, the model is still qualitatively representative even on shorter time scales than an interspike interval, and it is sufficiently accurate for many purposes, even on shorter time scales than may be expected due to the low-pass picture that motivated the model in the first place. For example, as we see in Fig.\,\ref{fig:sing_syn_event__vary_ib}(b), it is possible to design a receiver that accurately tracks the full temporal response to a single synapse event, indicating that this ``classical'' neurodynamic model is useful for representing discrete, spike-induced ``quantum'' effects. 

In general, it appears possible to design dendrites that are accurately represented by the phenemonological model to serve a variety of roles. If one wishes to have a receiver that responds rapidly to a single synapse event, the loop inductance should be relatively small. For a dendrite to play the role of a long-term integrator, the loop capacity should be large, and the time constant should be long. The phenomenological model can be accurate in either case, but the same dendrites should not be expected to serve both purposes.

\subsection{\label{sec:disynaptic_neuron}Disynaptic Neuron}
\begin{figure}[tbh]
\includegraphics[width=8.6cm]{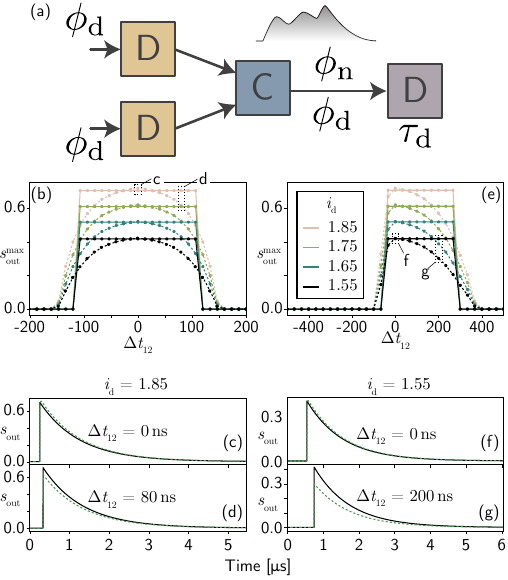}
\caption{\label{fig:coincidence_sequence}Coincidence and sequence detection. (a) Schematic of the neural circuit under consideration. (b-d) Coincidence detection. (b) Peak of the output dendrite response versus delay between two input synapse events. The solid lines were calculated with the spiking model, while the dashed lines result from the phenomenological model. (c, d) Example temporal responses for the two models. The black, solid line is the spiking model, and the green dashed line is the phenomenological model. (e-g) Sequence detection, wherein the two input synapses have different time constants. For both the coincidence and sequence calculations, the input dendrite was simulated with $i_\mathrm{b} = 1.55, 1.65, 1.75$, and $1.85$. The output dendrite had $i_\mathrm{b} = 1.85$. $\beta_\mathrm{in} = \beta_\mathrm{out} = 2\pi 10^2$. $\tau_\mathrm{in} = 250$\,ns, and $\tau_\mathrm{out} = 1$\,\textmu s.}
\end{figure}

The examples thus far have been of monosynaptic neurons connected to a single downstream dendrite. Much of the power of spiking neurons results from their ability to make use of information related to timing between input spikes on different synapses. To assess the ability of the phenomenological model to capture these effects, we now consider neurons with two input synapses, as shown schematically in Fig.\,\ref{fig:coincidence_sequence}(a), beginning with coincidence detection. We study a disynaptic neuron with synapses of the same time constant, $\tau_\mathrm{in} = 100$\,ns. Results are shown in Fig.\,\ref{fig:coincidence_sequence}. Figure \ref{fig:coincidence_sequence}(b) summarizes the data from the coincidence detector, plotting the peak of the downstream dendritic signal versus the relative delay between a synapse event on the first synapse and one on the second for several values of $i_\mathrm{d}$. The most striking feature of Fig.\,\ref{fig:coincidence_sequence}(b) is that the spiking model has a square response, while the phenomenological model is graded. These shapes occur because the spiking neuron communicates a binary response to the downstream dendrite: the synapse events are either close enough in time to drive the soma above threshold and produce a spike, or they are not. The same is not true for the phenomenological model, where analog information about the state of flux into the soma evokes a graded response in the downstream dendrite. Like the case of the low-pass-filtered spike train in Figs.\,\ref{fig:train_syn_events}(d) and (e), this is another instance where the downstream dendrite in the phenomenological model provides a better representation of the state of the soma because it has more information to work with, while the spiking neuron is seen to provide the best approximation possible to the full analog dendritic response given the restriction of only being able to communicate with binary spikes. While it may be appropriate to consider this to be a weakness or shortcoming of the phenomenological model as a tool to simulate spiking neurons, the time traces in Fig.\,\ref{fig:coincidence_sequence}(c) and (d) show that even when the phenomenological model departs from the spiking case, the agreement is still quantitatively decent and qualitatively excellent. As a tool for the study of network dynamics and algorithm development, the ability of the phenomenological model to capture coincidence detection supports the conjecture that a model without spikes has practical utility. Like the case of representing responses to single synapse events, this coincidence example confirms that the phenomenological model is not a reduction to rate coding, as times of individual spikes are not treated in rate codes.

The straightforward extension of the coincidence detector is to sequence detection using two input synapses on dendrites with different decay times, as shown in Fig.\,\ref{fig:coincidence_sequence}(e). Here we simulated $\tau_\mathrm{in 1} = 250$\,ns and $\tau_\mathrm{in 2} = 50$\,ns. The asymmetrical downstream dendrite response is shown in Fig.\,\ref{fig:coincidence_sequence}(e), with comparision time traces in Fig.\,\ref{fig:coincidence_sequence}(f) and (g). Just as in the case of coincidence detection, the spiking neuron has a binary response, while the phenomenological case produces a graded, analog response. Again, the time traces reveal reasonable qualitative accuracy, even at large time delays.

\section{\label{sec:classical_neurodynamics}Comparison to Classical Neurodynamics}
Reference \onlinecite{hopfield1986computing} introduced the terminology of ``classical neurodynamics'', which the authors used to refer to a point-neuron model in which spikes are not considered and only average activity is represented. The term ``classical'' was used to mean non-discrete, in contrast to ``quantum'' phenomena characterized by discrete energy levels, particles, or events. It is important to disambiguate that the words ``classical'' and ``quantum'' are used in this context only for analogy, and we are not discussing quantum neural networks in which superposition and entanglement are leveraged for computation. The paper by Hopfield and Tank included a conventional model of leaky-integrator point neurons in the rate-coding regime to represent classical neurodynamics. To compare the SOENs phenomenological model with classical neurodynamics, we refer to that model, where the equation of motion for the membrane potential on neuron $i$, $u_i$, is
\begin{equation}
\label{eq:classical_neurodynamics}
C_i\frac{du_i}{dt} = \sum_{j=1}^N T_{ij} f_j(u_j) - \frac{u_i}{R_i} + I_i.
\end{equation}
The nonlinear source term, $f_j(u_j)$, maps the membrane potential to an output firing rate. $C_i$ and $R_i$ are the membrane capacitance and resistance. $T_{ij}$ represents coupling, and $I_i$ is an external current drive. Equation \ref{eq:classical_neurodynamics} is also the first equation in the Hodgkin-Huxley model, with the other three equations capturing the dynamics of the gating variables that enter the coupling term. The analogy with SOENs follows from making the associations
\begin{equation}
\label{eq:classical_neurodynamics_to_soens}
\begin{split}
u_i &\rightarrow \phi_i, \\
f_i(u_i) &\rightarrow g(\phi_i,s_i).
\end{split}
\end{equation}

To make the correspondence between the two models clear, we need a differential equation describing the time evolution of $\phi_i$. Taking the time derivative of Eq.\,\ref{eq:main_equation__coupling}, using Eq.\,\ref{eq:main_equation}, and considering static inputs ($d\phi_i^\mathrm{ext}/dt = 0$), we have
\begin{equation}
\frac{d\phi_i}{dt} = \sum_j J_{ij}\gamma_j\,g(\phi_j,s_j) - \sum_j J_{ij}\frac{s_j}{\tau_j}.
\end{equation}
To move closer to the model of classical neurodynamics, we assume $\gamma_j = \gamma$, and $\tau_j = \tau$; we reduce the generality of the model by assuming these constants do not vary with $j$. This allows us to use Eq.\,\ref{eq:main_equation__coupling} to make the replacement $\sum_j J_{ij}\,s_j = \phi_i - \phi_i^\mathrm{ext}$, which gives
\begin{equation}
\label{eq:soens_neurodynamics}
\beta \frac{d\phi_i}{dt} = \sum_j J_{ij}\,g(\phi_j,s_j) - \alpha \phi_i + \alpha \phi_i^\mathrm{ext}.
\end{equation}
Recall that $\alpha = \beta/\tau$. Comparing Eq.\,\ref{eq:soens_neurodynamics} with Eq.\,\ref{eq:classical_neurodynamics}, we see the forms are nearly identical. The dimensionless inductance, $\beta$, plays the role of the capacitance, and the dimensionless resistance, $\alpha$, plays the role of the conductance. It is a relatively superficial difference that in SOENs these constants apply on the pre-synaptic side of the interaction, while in classical neurodynamics they apply on the post-synaptic side. They correspond when all such constants are taken to be equal. The more consequential difference is the fact that $s_i$ enters the source term, $g$. $s_i$ is a low-pass-filtered memory trace of the firing rate. The classical neurodynamic model has an output $V_i = f_i(u_i)$ which is not an input to $f_i$. Therefore, $f_i$ is a monotonically increasing function of only one input. In the case of SOENs with the simple dendrite comprising only a receiving and integrating loop \cite{shainline2023phenomenological}, $g(\phi_i,s_i)$ is a monotonically increasing function of the input $\phi_i$ (assuming total input flux is limited to $\Phi_0/2$, \cite{primavera2021active}), but it is a monotonically decreasing function of $s_i$, which means for constant or increasing $\phi$, the source term can have a positive or negative derivative. However, the main consequence of $s_i$ entering the source term is to cause the total response to saturate, leading to transfer functions that are close to sigmoidal, as was measured in Ref.\,\onlinecite{khan2022superconducting} (see also Fig.\,\ref{fig:transfer_functions} in Sec.\,\ref{sec:discussion} of the present work). The presence of $s$ in $g$ has the effect of shaping the response function into a nonlinear transfer function with a saturating nonlinearity. But the transfer function of classical neurodynamics, $f$, is typically taken to be exactly the same type of function, with a threshold and a saturating nonlinearity. Thus, the presence of $s$ in $g$ serves to render the analogy between the phenomenological SOENs model and classical neurodynamics quite close. In classical neurodynamics, $f$ is asserted to be a sigmoid, whereas in SOENs $g$ is made to have a similar shape through circuit engineering.

Considering how closely SOENs can map onto the classical neurodynamical model (Eq.\,\ref{eq:soens_neurodynamics} compared to Eq.\,\ref{eq:classical_neurodynamics}) and how well the SOENs phenomenological model can capture discrete events due to single spikes (Figs.\,\ref{fig:sing_syn_event} and Fig.\,\ref{fig:sing_syn_event__vary_ib}) and spike timing (Fig.\,\ref{fig:coincidence_sequence}), perhaps assuming the classical model forces us to neglect impulses of activity and timing between these impulses is giving up more than we must based on the mathematics. It is usually stated in words that $f$ represents the average firing rate of a neuron, which implies the model only applies to time-averaged spiking activity and cannot be defined to represent discrete firing events. But another interpretation is possible. As in the case of the $g$ term in SOENs, we can think of $f$ as a source term that adds charge to the neuronal membrane in a continuous and sometimes rapid fashion. As we have seen, this interpretation allows for the membrane potential to quickly respond to qualitatively and often quantitatively capture the response to a single spike input. A generating function with a sharper turn on than a sigmoid may be required from $f$ to capture this behavior, but that does not appreciably change the formulation of classical neurodynamics. 

Perhaps the major limitation of classical neurodynamics is not from the math itself, but rather the interpretation. Equation \ref{eq:classical_neurodynamics} is assumed to model neurons, when important computations can be achieved by interpreting this equation as a model of both dendrites and neurons, with possibly different generating functions employed for each, as is necessary for SOENs. If only neurons are considered, dendritic substructures of the full $\mathbf{J}$ weighted, directional, adjacency matrix are implicitly omitted, and we forfeit an important stage of processing complexity. To make the best use of temporal and spatial information across the network, it is not as important that we extend Eq.\,\ref{eq:classical_neurodynamics} to explicitly track discrete spikes as it is that we structure the network adjacency matrix, $\mathbf{J}$, and the dendritic time constants, $\tau$, heterogeneously across scales, giving each neuron an input dendritic tree and an output axonal arbor, just as we find in pyramidal neurons. Computational neuroscience literature abounds with models of active dendrites \cite{poirazi2020illuminating}; with SOENs we build them in artificial hardware, and their behavior is quite straightforward to model. With the partial phenomenological model, soma dynamics were still treated by modeling spikes. It was evident that dendrites followed the equations of classical neurodynamics, but only with the full phenomenological model presented here is it apparent that full superconducting optoelectronic neurons also map directly onto classical neurodynamics.
 
\section{\label{sec:discussion}Discussion}
The pragmatic motivation for this work was that the model, without reference to spikes, provides new avenues for numerical treatment. In this context, it is important that the phenomenological model successfully captures the qualitative features of the downstream dendritic response for a wide range of circuit parameters and neuronal stimuli, and in many cases the quantitative agreement is quite good. Appreciable additional work leveraging parallelized numerical implementation is required to determine exactly how much of a computational speedup results from this formalism. 

The model has important limitations. For example, entorhinal grid cells are critical in the formation cognitive maps. One model of their behavior involves on a grid cell responding to two input spiking neurons with spike rates that precess with an animal's velocity in different directions \cite{hasselmo2012we}. The grid cell then responds to coincidences between spikes from these two input neurons. As shown in Fig.\,\ref{fig:coincidence_sequence}, coincidences can be handled by the phenomenological model, but only if the spikes generated by the two upstream neurons are represented as discrete, highly localized temporal events. In the phenomenological framework, if these input neurons to the grid cell are deep in a complex network, they will not propagate discrete, highly localized events unless all upstream neurons and dendrites have short time constants to retain spike timing information. Accurate modeling of networks that produce grid cells is therefore one example of an instance in which a full spiking model is likely to provide benefits over the phenomenological approximation.

Given that the phenomenological model provides a qualitative and often quantitative representation of the dynamics of spiking neurons without any reference to spikes, and in some cases this model provides more accurate information about the state of the soma than is contained in the output spike train, we may ask the question: Why do neurons spike? There are many reasons \cite{brette2015philosophy, maass2015spike}. From a mathematical perspective, spikes are a unique basis set in that they can achieve translation and scale invariance, making spiking neural networks capable of representing a wide range of temporal and spatial phenomena, spanning length and time scales \cite{beer2020spiking}. Examples from the neuroscience cannon argue that spikes enable neurons to compute and communicate not just with rates but also precise coincidences \cite{konig1996integrator, koch2000role, spruston2008pyramidal}, sequences \cite{hawkins2016neurons}, in bursts \cite{naud2018sparse, zeldenrust2018neural}, and relative to a background phase  \cite{sirota2008entrainment} to achieve nested oscillations \cite{bonnefond2017communication} and rhythms \cite{wang2010neurophysiological, buzsaki2006rhythms} that direct attention \cite{fries2015rhythms}, and enable rapidly adaptive processing through multistability \cite{deco2016metastability}. Across these examples, the computations performed by dendrites play a crucial role. Spikes have low latency and timing jitter \cite{vanrullen2005spike}, particularly when the stimuli require precision timing \cite{mainen1995reliability}. They provide a unique means to send information accurately, rapidly, and over long distances, all of which enable highly interconnected networks to exchange information in a robust manner, which equips organisms for survival in a variety of circumstances \cite{humphries2021spike}. When stimulus changes rapidly, neurons can represent that change rapidly with spikes. When understanding a stimulus requires precise timing---such as in the auditory cortex where timing differences between two cochlea map to spatial location of a source---spikes are excellent for representing that precise timing. Yet in different circumstances or in different brain regions when stimuli are slowly varying, spikes can still do an excellent job of representing these stimuli as well. It is possible that spikes are important for all the above reasons related to energy, computation, and communication, but that spikes are still not the most important quantity for the network designer to track in their model.

One need not choose globally between a spike-timing representation of information or a slower, rate-based representation. Spikes can do both, and the model presented here can capture both modalities. In typical modeling formalisms, the observables are either delta-function spike trains or time-averaged spike rates. With this formalism for SOENs, we dispense with both; the observables are the signals in all dendrites of the network. The vector $\mathbf{s}$ uniquely identifies a point in configuration space, and its temporal evolution, Eq.\,\ref{eq:main_equation__vector_form}, provides a complete description of the system. Sometimes $\mathbf{s}$ varies quickly, resembling spikes. Other times it varies slowly, resembling rates. In all cases, the neurons are adapting dynamically in response to the stimulus. This picture is consistent with the perspective from a dendrite's point of view. At any moment, a dendrite cannot convey the rate of afferent synapse events it has recently experienced, nor can it report the precise time of the most recent or any other input spike. It can only communicate the signal it has stored in its integration loop. With properly configured dendritic arbors and neural network graph structure, these signals are sufficient to represent the input, whether that input is changing rapidly or slowly. The phrase ``properly configured'' refers to the choices of all elements of the $\mathbf{J}$ matrix, the loop-filling rates $\boldsymbol{\gamma}$, and the time constants, $\boldsymbol{\tau}$, all of which must be chosen based on prior information regarding the statistical structure of the input stimuli the network is designed to represent.

In the specific case of SOENs, we conjecture that the objective of a soma is to communicate information about its state to its downstream synapses. From the physical arguments detailed in Refs.\,\onlinecite{shainline2021optoelectronic} and \onlinecite{shainline2019superconducting} we conclude the optimal medium to communicate this information is light at the few-photon level. But this could be done in at least two ways. One way is to keep a light source on continuously, and to have the luminosity of that source (set by the current through the diode) be an analog function of the state of the soma. The downstream synapses would then receive a signal that, on average, reflected the state of the soma, and the time variability due to the Poisson statistics of the diode would be a source of noise. Another way is to use spikes. The soma produces discrete, short pulses at times that indicate it crossed a threshold, and these spikes include enough photons for every recipient synapse to receive at least one photon with low timing jitter. In this case noise is reduced. The spike case improves over the analog Poisson case because the uncertainty faced by the synapse regarding the state of the soma is more rapidly reduced when the soma state changes significantly, while the Poisson source must be integrated over a longer time to reduce noise from timing jitter and gain confidence (in a Bayesian sense) that the varying photon flux has indeed changed, representing change in the soma's state. The Poisson source could be made less variable, but only by increasing the photon flux, which costs energy. Thus, in the specific case of SOENs, where faint-light signals are received by single-photon detectors, several arguments can be made in favor of the energy and temporal benefits of spikes relative to an alternative analog approach. Comparing to biology, we find similar arguments can be made. If each soma had to maintain its entire axonal arbor at a specified analog voltage that varied in time to track the state of the soma's membrane, significantly more energy would be required, or the signal would be subject to appreciably more noise. 

These arguments recapitulate what has already been presented in the literature regarding the reasons neurons spike. But the model presented here does offer a new lens through which to view the situation. In Figs.\,\ref{fig:train_syn_events}(d), (e) and \ref{fig:coincidence_sequence} we showed examples where the phenomenological model propagated analog information about the state of the soma to the downstream dendrites, which allow the phenomenological model to provide a better representation of the state of the soma than was achieved through discrete, binary spikes. In that context we lamented the inability of the phenomenological model to quantitatively match the saw-tooth response of the full spiking neuron, but we can turn the picture around. Instead, one could conjecture that a neuron wishes to, or ideally would, cast a time-continuous, analog signal to all downstream synapses so they could have accurate information about the state of the soma at all times. But no neuron---biological or technological---can accomplish this due to physical constraints. Such signaling is practically difficult and energetically expensive, so neurons resort to spikes as efficient, and quick approximations to the full analog, time-continuous state of the soma. We can then see the message of Figs.\,\ref{fig:train_syn_events}(d), (e) and \ref{fig:coincidence_sequence} inversely: spiking neurons do a pretty good job of approximating the signal the soma would like to send, but their physical limitations force them to approximate the desired smooth signals with jagged responses. Low-pass filtering will smooth the time series, and during realistic network operation, the square corners are rounded by various noise mechanisms, so expending additional energy to capture the analog curve is not a beneficial endeavor.

An additional result of the model is conceptual simplification. We can consider systems entirely in terms of the state vector $\mathbf{s}$ at all levels of the network, and each $s_i$ is defined at all times purely in terms of other values of $s_j$ and the fixed $\mathbf{J}$ matrix. The pulsatile body of a spiking neuron and the recipient synapse have both been eliminated from the picture, and what remains is a network of interconnected dendrites, performing summation, temporal integration, and nonlinear transfer functions on their inputs. Everything is continuous in time and constructed from the same dendritic building blocks, which are simple yet versatile. Even the learning circuits which establish plasticity signals are governed by the same picture and the same differential equations. Yet eliminating the spiking somas and recipient synapses in the model does not require forfeiting their benefits in hardware. Superconducting optoelectronic neurons still spike, which means their transmitters are usually off, saving appreciable energy. These spikes can travel far and fast with low noise, updating downstream knowledge at light speed, in contrast to what would be achieved if an analog electrical signal were required to charge a large axonal arbor. And dendrites can still make use of timing information for coincidence and sequence detection, phase and burst coding, and spike-timing-dependent learning protocols. 

It may be the case that the first phenomenological model of Ref.\,\onlinecite{shainline2023phenomenological}, which included spikes, produces results that are sensitive to the precise timing of spikes, even in low-pass-filtered cases, such as when the peaks of the saw-tooth response of Fig.\,\ref{fig:step_response} affect downstream behavior, with discrepancies observed based on few-nanosecond timing jitter of synapse events, a temporal resolution at which we should not expect the present model to be accurate. Dendritic and neuronal nonlinearities make it likely that the response across trials could be chaotic with respect to minor variations in such timing. One must then consider the timing jitter of physically instantiated components. The dominant contribution to the jitter of synapses events is likely to result from stochasticity in photon production times due to the light source. A one-nanosecond, direct-gap emitter would be ideal, but there is no guarantee such a source will become viable at scale with low cost when integrated with the rest of the SOENs fabrication process. A silicon light source is another possibility, in which case the emission time constant is likely to be between 30\,ns and 100\,ns, with commensurate uncertainty in timing of synapse events. It is necessary in such cases that any computation or algorithm implemented by the network be resilient to such timing jitter. One approach to achieve this in modeling is to run Monte Carlo trials, inject the expected jitter into the simulation, and average over network outcomes. The average of this procedure will lead to a result comparable to the output of the framework presented here.

\begin{figure}[tbh]
\includegraphics[width=8.6cm]{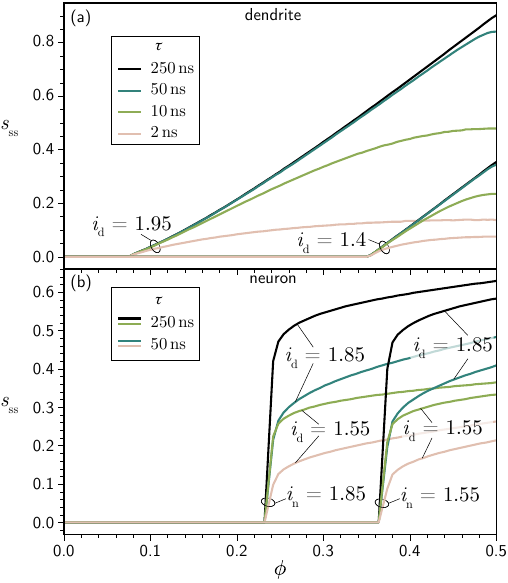}
\caption{\label{fig:transfer_functions}Dendrite and neuron steady-state transfer functions obtained through solution of Eq.\,\ref{eq:steady_state}. (a) The case of a dendrite with $g_\mathrm{d}$ as shown in Fig.\,\ref{fig:source_functions}(a) and (b). (b) The case of a neuron coupled to a downstream synapse and dendrite with $g_\mathrm{n}$ as shown in Fig.\,\ref{fig:source_functions}(c)-(f).}
\end{figure}

The phenomenological model offers a transparent framework for formal analysis and renders various learning modalities and training algorithms applicable to SOENs. We showed above that the analogy between SOENs viewed through this lens and classical neurodynamics is tight. The body of knowledge generated in the context of classical neurodynamics therefore applies in bulk to SOENs as well. Several useful machine-learning algorithms applied to neural networks are motivated by the basic picture of neurons gleaned from classical neurodynamics. For example, backpropagation \cite{lecun2015deep, goodfellow2016deep} has been the most impactful technique for credit assignment in artificial intelligence, and the transfer functions used in that formalism are typically interpreted as representing average neuron or population spike rates. Yet it has been difficult to implement backpropagation in spiking neural networks due to the discrete, non-differentiable nature of spike signals. In the case of the SOENs phenomenological model, all signals are continuous in time, and the model can be used to calculate steady-state dendritic transfer functions and their derivatives. The steady state of the model can be obtained from Eq.\,\ref{eq:main_equation},
\begin{equation}
\label{eq:steady_state}
s_\mathrm{ss}(\phi_\mathrm{ss}) = \frac{1}{\alpha}\;g(\phi_\mathrm{ss},s_\mathrm{ss}),
\end{equation}
an equation that can be straightforwardly solved self-consistently, providing the continuous, differentiable transfer functions shown in Fig.\,\ref{fig:transfer_functions}.

The phenomenological model for dendrites was derived directly from a Lagrangian and the Euler-Lagrange equations that result from minimizing action. An approximation was made to treat the JJs as a voltage source of a particular form, and the quantitative agreement between the phenomenological and circuit models indicates the purest Lagrangian of the system is well represented by the approximate form. We can proceed as if the course of dendrites through configuration space follows exactly the path of least action. The phenomenological model for neurons was conjectured on conceptual grounds by analogy to the dendritic case on slower time scales. A first-principles derivation from a Lagrangian is more cumbersome for a neuron due to the complexities of the superconducting thin-film amplifiers and semiconductor transmitter circuits involved. But we have shown that the full circuit equations and the much simpler Lagrangian representation---again a simple circuit with a peculiar but tractable voltage source---provides a decent representation of the dynamics. At the scale of neurons, too, we can proceed as if the network trajectory through configuration space closely follows the path of least action of the much simpler phenomenological circuit of Fig.\,\ref{fig:circuits}(c). Complete evolution of the network follows from foundational physical principles. The tools of statistical mechanics can be applied. 

The framework developed here may form a bridge to energy models \cite{lecun2006tutorial, bengio2017stdp, hoover2024energy} such as Hopfield networks \cite{hopfield1982neural, hopfield1984neurons, ramsauer2020hopfield, krotov2020large} and associated training algorithms such as equilibrium propagation \cite{scellier2017equilibrium, mansingh2024robust}, wherein an energy is associated with any state of the network, and the task of training is to find update rules that take an error or cost function and use this information to change the parameters of the network in such a way that states of low energy are also states of low error. The equation of motion for classical neurodynamics, Eq.\,\ref{eq:classical_neurodynamics}, was used in Ref.\,\onlinecite{yan2013nonequilibrium} as a starting point to formulate a nonequilibrium statistical mechanical model of neural networks. Because the dynamical equation of SOENs under the approximations considered here match classical neurodynamics, the formalism of Ref.\,\onlinecite{yan2013nonequilibrium} applies to SOENs as well. The Lagrangian model presented here can be straightforwardly extended to an energy function. The energy in a given dendrite is
\begin{equation}
\label{eq:energy}
E_i = \frac{1}{2}\beta s_i^2.
\end{equation}
Given the threshold-linear transfer function of a dendrite, Fig.\,\ref{fig:transfer_functions}(a), which also happens to be symmetric about $\phi = 0$, a dendrite with two inputs of equal magnitude and opposite sign has energy $E \sim (s_1-s_2)^2$, which is a commonly used cost function for energy models. Therefore, in SOENs, the physical energy stored in a dendrite is also the ubiquitous energy used as an error function, a fact with immediate ramifications for predictive coding \cite{rao1999predictive} and the free-energy principle \cite{friston2010free}. With SOENs, it is trivial to engineer dendrites or neurons whose activity ceases only when top-down and bottom-up signals match.

In the energy picture, the parameters $J_{ij}$ affect the energy landscape and so can be chosen to sculpt a given network to the statistical signatures of a certain data set. Additional plasticity dendrites can also be added to dynamically sculpt the landscape, enabling phenomena such as multistability \cite{mazzucato2015dynamics, la2019cortical}, winnerless competition \cite{rabinovich2001dynamical, seliger2003dynamics, rabinovich2008transient} and episodic memories \cite{karuvally2022energy}. To implement biologically inspired learning techniques, such as Hebbian update rules \cite{sejnowski1999book}, spike-timing-dependent plasticity \cite{markram2012spike}, and three-factor update \cite{joel2002actor, fremaux2013reinforcement, fremaux2016neuromodulated, kusmierz2017learning, gerstner2018eligibility} (relevant for reinforcement learning), the fact that the model captures coincidence and sequence events is essential, and the values of $s$ present at all dendrites can be used for credit assignment and update of the entire dendritic arbor in the presence of a top-down reward or error signal. These examples point to a plethora of possible future investigations. The value of the model presented here will be gauged by its numerical speed and its ability to bridge SOENs to these vast bodies of other work so this promising hardware can be made practically useful.

\section{\label{sec:acknowledgements}Acknowledgements}
This is a contribution of NIST, an agency of the US government, not subject to copyright.

\bibliographystyle{unsrt}
\bibliography{relating_soens_neurodynamics}

\begin{thebibliography}{10}

\bibitem{shainline2023phenomenological}
Jeffrey~M Shainline, Bryce~A Primavera, and Saeed Khan.
\newblock Phenomenological model of superconducting optoelectronic loop
  neurons.
\newblock {\em Physical Review Research}, 5(1):013164, 2023.

\bibitem{nawrocki2016mini}
Robert~A Nawrocki, Richard~M Voyles, and Sean~E Shaheen.
\newblock A mini review of neuromorphic architectures and implementations.
\newblock {\em IEEE Transactions on Electron Devices}, 63(10):3819--3829, 2016.

\bibitem{furber2016large}
Steve Furber.
\newblock Large-scale neuromorphic computing systems.
\newblock {\em Journal of neural engineering}, 13(5):051001, 2016.

\bibitem{schuman2017survey}
Catherine~D Schuman, Thomas~E Potok, Robert~M Patton, J~Douglas Birdwell,
  Mark~E Dean, Garrett~S Rose, and James~S Plank.
\newblock A survey of neuromorphic computing and neural networks in hardware.
\newblock {\em arXiv preprint arXiv:1705.06963}, 2017.

\bibitem{shainline2018circuit}
Jeffrey~M Shainline, Sonia~M Buckley, Adam~N McCaughan, Jeff Chiles, Amir
  Jafari-Salim, Richard~P Mirin, and Sae~Woo Nam.
\newblock Circuit designs for superconducting optoelectronic loop neurons.
\newblock {\em Journal of {A}pplied {P}hysics}, 124(15):152130, 2018.

\bibitem{shainline2021optoelectronic}
Jeffrey~M Shainline.
\newblock Optoelectronic intelligence.
\newblock {\em Applied {P}hysics {L}etters}, 118(16):160501, 2021.

\bibitem{buckley2017all}
Sonia Buckley, Jeffrey Chiles, Adam~N McCaughan, Galan Moody, Kevin~L
  Silverman, Martin~J Stevens, Richard~P Mirin, Sae~Woo Nam, and Jeffrey~M
  Shainline.
\newblock All-silicon light-emitting diodes waveguide-integrated with
  superconducting single-photon detectors.
\newblock {\em Applied {P}hysics {L}etters}, 111(14):141101, 2017.

\bibitem{chiles2017multi}
Jeff Chiles, Sonia Buckley, Nima Nader, Sae~Woo Nam, Richard~P Mirin, and
  Jeffrey~M Shainline.
\newblock Multi-planar amorphous silicon photonics with compact interplanar
  couplers, cross talk mitigation, and low crossing loss.
\newblock {\em {APL} {P}hotonics}, 2(11):116101, 2017.

\bibitem{chiles2018design}
Jeff Chiles, Sonia~M Buckley, Sae~Woo Nam, Richard~P Mirin, and Jeffrey~M
  Shainline.
\newblock Design, fabrication, and metrology of 10$\times$ 100 multi-planar
  integrated photonic routing manifolds for neural networks.
\newblock {\em {APL} {P}hotonics}, 3(10):106101, 2018.

\bibitem{mccaughan2019superconducting}
Adam~N McCaughan, Varun~B Verma, Sonia~M Buckley, JP~Allmaras, AG~Kozorezov,
  AN~Tait, SW~Nam, and JM~Shainline.
\newblock A superconducting thermal switch with ultrahigh impedance for
  interfacing superconductors to semiconductors.
\newblock {\em Nature {E}lectronics}, 2(10):451--456, 2019.

\bibitem{khan2022superconducting}
Saeed Khan, Bryce~A Primavera, Jeff Chiles, Adam~N McCaughan, Sonia~M Buckley,
  Alexander~N Tait, Adriana Lita, John Biesecker, Anna Fox, David Olaya, et~al.
\newblock Superconducting optoelectronic single-photon synapses.
\newblock {\em Nature {E}lectronics}, 5:650--659, 2022.

\bibitem{primavera2023programmable}
Bryce~A Primavera, Saeed Khan, Richard~P Mirin, Sae~Woo Nam, and Jeffrey~M
  Shainline.
\newblock Programmable superconducting optoelectronic single-photon synapses
  with integrated multi-state memory.
\newblock {\em arXiv preprint arXiv:2311.05881}, 2023.

\bibitem{hodgkin1952quantitative}
Alan~L Hodgkin and Andrew~F Huxley.
\newblock A quantitative description of membrane current and its application to
  conduction and excitation in nerve.
\newblock {\em The Journal of physiology}, 117(4):500, 1952.

\bibitem{gerstner2002spiking}
W.~Gerstner and W.M. Kistler.
\newblock {\em Spiking neuron models: Single neurons, populations, plasticity}.
\newblock Cambridge university press, 2002.

\bibitem{van1998principles}
T.~Van Duzer and C.W. Turner.
\newblock {\em Principles of superconductive devices and circuits}.
\newblock Prentice {H}all, USA, second edition, 1998.

\bibitem{kadin1999introduction}
Alan~M Kadin.
\newblock {\em Introduction to superconducting circuits}.
\newblock Wiley-{I}nterscience, 1999.

\bibitem{tinkham2004introduction}
Michael Tinkham.
\newblock {\em Introduction to superconductivity}.
\newblock Dover, 2004.

\bibitem{clarke2006squid}
J.~Clarke and A.I. Braginski, editors.
\newblock {\em The SQUID handbook}.
\newblock Wiley-VCH.

\bibitem{crotty2010josephson}
Patrick Crotty, Dan Schult, and Ken Segall.
\newblock Josephson junction simulation of neurons.
\newblock {\em Physical Review E}, 82(1):011914, 2010.

\bibitem{hopfield1986computing}
John~J Hopfield and David~W Tank.
\newblock Computing with neural circuits: A model.
\newblock {\em Science}, 233(4764):625--633, 1986.

\bibitem{brette2015philosophy}
Romain Brette.
\newblock Philosophy of the spike: rate-based vs. spike-based theories of the
  brain.
\newblock {\em Frontiers in systems neuroscience}, 9:151, 2015.

\bibitem{wells1938application}
DA~Wells.
\newblock Application of the {L}agrangian equations to electrical circuits.
\newblock {\em Journal of Applied Physics}, 9(5):312--320, 1938.

\bibitem{goldstein2002classical}
Herbert Goldstein, Charles Poole, and John Safko.
\newblock {\em Classical mechanics}.
\newblock American Association of Physics Teachers, 2002.

\bibitem{alahmadi2020performance}
Sarah AlAhmadi, Thaha Mohammed, Aiiad Albeshri, Iyad Katib, and Rashid Mehmood.
\newblock Performance analysis of sparse matrix-vector multiplication ({SpMV})
  on graphics processing units ({GPUs}).
\newblock {\em Electronics}, 9(10):1675, 2020.

\bibitem{primavera2021active}
Bryce~A Primavera and Jeffrey~M Shainline.
\newblock An active dendritic tree can mitigate fan-in limitations in
  superconducting neurons.
\newblock {\em Applied Physics Letters}, 119(24), 2021.

\bibitem{poirazi2020illuminating}
Panayiota Poirazi and Athanasia Papoutsi.
\newblock Illuminating dendritic function with computational models.
\newblock {\em Nature Reviews Neuroscience}, 21(6):303--321, 2020.

\bibitem{hasselmo2012we}
Michael~E Hasselmo.
\newblock {\em How we remember: Brain mechanisms of episodic memory}.
\newblock MIT press, 2012.

\bibitem{maass2015spike}
Wolfgang Maass.
\newblock To spike or not to spike: that is the question.
\newblock {\em Proceedings of the IEEE}, 103(12):2219--2224, 2015.

\bibitem{beer2020spiking}
Michael Beer, Julio Urenda, Olga Kosheleva, and Vladik Kreinovich.
\newblock Why spiking neural networks are efficient: A theorem.
\newblock In {\em International Conference on Information Processing and
  Management of Uncertainty in Knowledge-Based Systems}, pages 59--69.
  Springer, 2020.

\bibitem{konig1996integrator}
Peter K{\"o}nig, Andreas~K Engel, and Wolf Singer.
\newblock Integrator or coincidence detector? the role of the cortical neuron
  revisited.
\newblock {\em Trends in neurosciences}, 19(4):130--137, 1996.

\bibitem{koch2000role}
Christof Koch and Idan Segev.
\newblock The role of single neurons in information processing.
\newblock {\em Nature neuroscience}, 3(11):1171--1177, 2000.

\bibitem{spruston2008pyramidal}
Nelson Spruston.
\newblock Pyramidal neurons: dendritic structure and synaptic integration.
\newblock {\em Nature Reviews Neuroscience}, 9(3):206--221, 2008.

\bibitem{hawkins2016neurons}
Jeff Hawkins and Subutai Ahmad.
\newblock Why neurons have thousands of synapses, a theory of sequence memory
  in neocortex.
\newblock {\em Frontiers in neural circuits}, 10:174222, 2016.

\bibitem{naud2018sparse}
Richard Naud and Henning Sprekeler.
\newblock Sparse bursts optimize information transmission in a multiplexed
  neural code.
\newblock {\em Proceedings of the National Academy of Sciences},
  115(27):E6329--E6338, 2018.

\bibitem{zeldenrust2018neural}
Fleur Zeldenrust, Wytse~J Wadman, and Bernhard Englitz.
\newblock Neural coding with bursts—current state and future perspectives.
\newblock {\em Frontiers in computational neuroscience}, 12:48, 2018.

\bibitem{sirota2008entrainment}
Anton Sirota, Sean Montgomery, Shigeyoshi Fujisawa, Yoshikazu Isomura, Michael
  Zugaro, and Gy{\"o}rgy Buzs{\'a}ki.
\newblock Entrainment of neocortical neurons and gamma oscillations by the
  hippocampal theta rhythm.
\newblock {\em Neuron}, 60(4):683--697, 2008.

\bibitem{bonnefond2017communication}
Mathilde Bonnefond, Sabine Kastner, and Ole Jensen.
\newblock Communication between brain areas based on nested oscillations.
\newblock {\em eneuro}, 4(2), 2017.

\bibitem{wang2010neurophysiological}
Xiao-Jing Wang.
\newblock Neurophysiological and computational principles of cortical rhythms
  in cognition.
\newblock {\em Physiological reviews}, 90(3):1195--1268, 2010.

\bibitem{buzsaki2006rhythms}
Gy{\"o}rgy Buzs{\'a}ki.
\newblock {\em Rhythms of the Brain}.
\newblock Oxford university press, 2006.

\bibitem{fries2015rhythms}
Pascal Fries.
\newblock Rhythms for cognition: communication through coherence.
\newblock {\em Neuron}, 88(1):220--235, 2015.

\bibitem{deco2016metastability}
Gustavo Deco and Morten~L Kringelbach.
\newblock Metastability and coherence: extending the communication through
  coherence hypothesis using a whole-brain computational perspective.
\newblock {\em Trends in neurosciences}, 39(3):125--135, 2016.

\bibitem{vanrullen2005spike}
Rufin VanRullen, Rudy Guyonneau, and Simon~J Thorpe.
\newblock Spike times make sense.
\newblock {\em Trends in neurosciences}, 28(1):1--4, 2005.

\bibitem{mainen1995reliability}
Zachary~F Mainen and Terrence~J Sejnowski.
\newblock Reliability of spike timing in neocortical neurons.
\newblock {\em Science}, 268(5216):1503--1506, 1995.

\bibitem{humphries2021spike}
Mark Humphries.
\newblock {\em The spike: An epic journey through the brain in 2.1 seconds}.
\newblock Princeton University Press, 2021.

\bibitem{shainline2019superconducting}
Jeffrey~M Shainline, Sonia~M Buckley, Adam~N McCaughan, Jeffrey~T Chiles, Amir
  Jafari~Salim, Manuel Castellanos-Beltran, Christine~A Donnelly, Michael~L
  Schneider, Richard~P Mirin, and Sae~Woo Nam.
\newblock Superconducting optoelectronic loop neurons.
\newblock {\em Journal of Applied Physics}, 126(4), 2019.

\bibitem{lecun2015deep}
Yann LeCun, Yoshua Bengio, and Geoffrey Hinton.
\newblock Deep learning.
\newblock {\em nature}, 521(7553):436--444, 2015.

\bibitem{goodfellow2016deep}
Ian Goodfellow, Yoshua Bengio, and Aaron Courville.
\newblock {\em Deep learning}.
\newblock MIT press, 2016.

\bibitem{lecun2006tutorial}
Yann LeCun, Sumit Chopra, Raia Hadsell, M~Ranzato, and Fujie Huang.
\newblock A tutorial on energy-based learning.
\newblock {\em Predicting structured data}, 1(0), 2006.

\bibitem{bengio2017stdp}
Yoshua Bengio, Thomas Mesnard, Asja Fischer, Saizheng Zhang, and Yuhuai Wu.
\newblock Stdp-compatible approximation of backpropagation in an energy-based
  model.
\newblock {\em Neural computation}, 29(3):555--577, 2017.

\bibitem{hoover2024energy}
Benjamin Hoover, Yuchen Liang, Bao Pham, Rameswar Panda, Hendrik Strobelt,
  Duen~Horng Chau, Mohammed Zaki, and Dmitry Krotov.
\newblock Energy transformer.
\newblock {\em Advances in Neural Information Processing Systems}, 36, 2024.

\bibitem{hopfield1982neural}
John~J Hopfield.
\newblock Neural networks and physical systems with emergent collective
  computational abilities.
\newblock {\em Proceedings of the national academy of sciences},
  79(8):2554--2558, 1982.

\bibitem{hopfield1984neurons}
John~J Hopfield.
\newblock Neurons with graded response have collective computational properties
  like those of two-state neurons.
\newblock {\em Proceedings of the national academy of sciences},
  81(10):3088--3092, 1984.

\bibitem{ramsauer2020hopfield}
Hubert Ramsauer, Bernhard Sch{\"a}fl, Johannes Lehner, Philipp Seidl, Michael
  Widrich, Thomas Adler, Lukas Gruber, Markus Holzleitner, Milena Pavlovi{\'c},
  Geir~Kjetil Sandve, et~al.
\newblock Hopfield networks is all you need.
\newblock {\em arXiv preprint arXiv:2008.02217}, 2020.

\bibitem{krotov2020large}
Dmitry Krotov and John Hopfield.
\newblock Large associative memory problem in neurobiology and machine
  learning.
\newblock {\em arXiv preprint arXiv:2008.06996}, 2020.

\bibitem{scellier2017equilibrium}
Benjamin Scellier and Yoshua Bengio.
\newblock Equilibrium propagation: Bridging the gap between energy-based models
  and backpropagation.
\newblock {\em Frontiers in computational neuroscience}, 11:24, 2017.

\bibitem{mansingh2024robust}
Siddharth Mansingh, Michal Kucer, Garrett Kenyon, Juston Moore, and Michael
  Teti.
\newblock How robust are energy-based models trained with equilibrium
  propagation?
\newblock {\em arXiv preprint arXiv:2401.11543}, 2024.

\bibitem{yan2013nonequilibrium}
Han Yan, Lei Zhao, Liang Hu, Xidi Wang, Erkang Wang, and Jin Wang.
\newblock Nonequilibrium landscape theory of neural networks.
\newblock {\em Proceedings of the National Academy of Sciences},
  110(45):E4185--E4194, 2013.

\bibitem{rao1999predictive}
Rajesh~PN Rao and Dana~H Ballard.
\newblock Predictive coding in the visual cortex: a functional interpretation
  of some extra-classical receptive-field effects.
\newblock {\em Nature neuroscience}, 2(1):79--87, 1999.

\bibitem{friston2010free}
Karl Friston.
\newblock The free-energy principle: a unified brain theory?
\newblock {\em Nature reviews neuroscience}, 11(2):127--138, 2010.

\bibitem{mazzucato2015dynamics}
Luca Mazzucato, Alfredo Fontanini, and Giancarlo La~Camera.
\newblock Dynamics of multistable states during ongoing and evoked cortical
  activity.
\newblock {\em Journal of Neuroscience}, 35(21):8214--8231, 2015.

\bibitem{la2019cortical}
Giancarlo La~Camera, Alfredo Fontanini, and Luca Mazzucato.
\newblock Cortical computations via metastable activity.
\newblock {\em Current opinion in neurobiology}, 58:37--45, 2019.

\bibitem{rabinovich2001dynamical}
M~Rabinovich, A~Volkovskii, P~Lecanda, R~Huerta, HDI Abarbanel, and G~Laurent.
\newblock Dynamical encoding by networks of competing neuron groups: winnerless
  competition.
\newblock {\em Physical review letters}, 87(6):068102, 2001.

\bibitem{seliger2003dynamics}
Philip Seliger, Lev~S Tsimring, and Mikhail~I Rabinovich.
\newblock Dynamics-based sequential memory: winnerless competition of patterns.
\newblock {\em Physical Review E}, 67(1):011905, 2003.

\bibitem{rabinovich2008transient}
Misha Rabinovich, Ramon Huerta, and Gilles Laurent.
\newblock Transient dynamics for neural processing.
\newblock {\em Science}, 321(5885):48--50, 2008.

\bibitem{karuvally2022energy}
Arjun Karuvally, Terry~J Sejnowski, and Hava~T Siegelmann.
\newblock Energy-based general sequential episodic memory networks at the
  adiabatic limit.
\newblock {\em arXiv preprint arXiv:2212.05563}, 2022.

\bibitem{sejnowski1999book}
Terrence~J Sejnowski.
\newblock The book of {H}ebb.
\newblock {\em Neuron}, 24(4):773--776, 1999.

\bibitem{markram2012spike}
Henry Markram, Wulfram Gerstner, and Per~Jesper Sj{\"o}str{\"o}m.
\newblock Spike-timing-dependent plasticity: a comprehensive overview.
\newblock {\em Frontiers in synaptic neuroscience}, 4:2, 2012.

\bibitem{joel2002actor}
Daphna Joel, Yael Niv, and Eytan Ruppin.
\newblock Actor--critic models of the basal ganglia: New anatomical and
  computational perspectives.
\newblock {\em Neural networks}, 15(4-6):535--547, 2002.

\bibitem{fremaux2013reinforcement}
Nicolas Fr{\'e}maux, Henning Sprekeler, and Wulfram Gerstner.
\newblock Reinforcement learning using a continuous time actor-critic framework
  with spiking neurons.
\newblock {\em PLoS computational biology}, 9(4):e1003024, 2013.

\bibitem{fremaux2016neuromodulated}
Nicolas Fr{\'e}maux and Wulfram Gerstner.
\newblock Neuromodulated spike-timing-dependent plasticity, and theory of
  three-factor learning rules.
\newblock {\em Frontiers in neural circuits}, 9:85, 2016.

\bibitem{kusmierz2017learning}
{\L}ukasz Ku{\'s}mierz, Takuya Isomura, and Taro Toyoizumi.
\newblock Learning with three factors: modulating hebbian plasticity with
  errors.
\newblock {\em Current opinion in neurobiology}, 46:170--177, 2017.

\bibitem{gerstner2018eligibility}
Wulfram Gerstner, Marco Lehmann, Vasiliki Liakoni, Dane Corneil, and Johanni
  Brea.
\newblock Eligibility traces and plasticity on behavioral time scales:
  experimental support of neohebbian three-factor learning rules.
\newblock {\em Frontiers in neural circuits}, 12:53, 2018.

\end{thebibliography}

\end{document}